\documentclass{article}
\usepackage{ijcai25}

\usepackage{times}
\usepackage{soul}
\usepackage{url}
\usepackage[hidelinks]{hyperref}
\usepackage[utf8]{inputenc}
\usepackage[small]{caption}
\usepackage{graphicx}
\usepackage{amsmath}
\usepackage{amsthm}
\usepackage{booktabs}
\usepackage{algorithm}
\usepackage{algorithmic}
\usepackage[switch]{lineno}

\usepackage{amssymb}
\usepackage{bm}
\usepackage{tcolorbox}
\usepackage{lipsum} 
\usepackage{enumitem}
\usepackage{multirow}

\title{Language-Based Bayesian Optimization Research Assistant (BORA)\footnote{This work is accepted and will be part of the 34th International Joint Conference on Artificial Intelligence (IJCAI) proceedings.}}

\author {
    Abdoulatif Cissé$^{1,2}$\and
    Xenophon Evangelopoulos$^{1,2}$\and
    Vladimir V. Gusev$^3$\And
    Andrew I. Cooper$^{1,2}$\\
\affiliations
$^1$Department of Chemistry, University of Liverpool, England, UK\\
$^2$Leverhulme Research Centre for Functional Materials Design, University of Liverpool, England, UK\\
$^3$Department of Computer Science, University of Liverpool, England, UK\\
\emails
\{abdoulatif.cisse, evangx, vladimir.gusev, aicooper\}@liverpool.ac.uk
}

\begin{document}

\maketitle

\begin{abstract}
Many important scientific problems involve multivariate optimization coupled with slow and laborious experimental measurements. These high-dimensional searches can be defined by complex, non-convex optimization landscapes that resemble needle-in-a-haystack surfaces, leading to entrapment in local minima. Contextualizing optimizers with human domain knowledge is a powerful approach to guide searches to localized fruitful regions. However, this approach is susceptible to human confirmation bias. It is also challenging for domain experts to keep track of the rapidly expanding scientific literature. Here, we propose the use of Large Language Models (LLMs) for contextualizing Bayesian optimization (BO) via a hybrid optimization framework that intelligently and economically blends stochastic inference with domain knowledge-based insights from the LLM, which is used to suggest new, better-performing areas of the search space for exploration. Our method fosters user engagement by offering real-time commentary on the optimization progress, explaining the reasoning behind the search strategies. We validate the effectiveness of our approach on synthetic benchmarks with up to 15 variables and demonstrate the ability of LLMs to reason in four real-world experimental tasks where context-aware suggestions boost optimization performance substantially.
\end{abstract}

\section{Introduction}
\label{sec:intro}
\begin{figure}[t]
    \centering
    \includegraphics[width=\columnwidth]{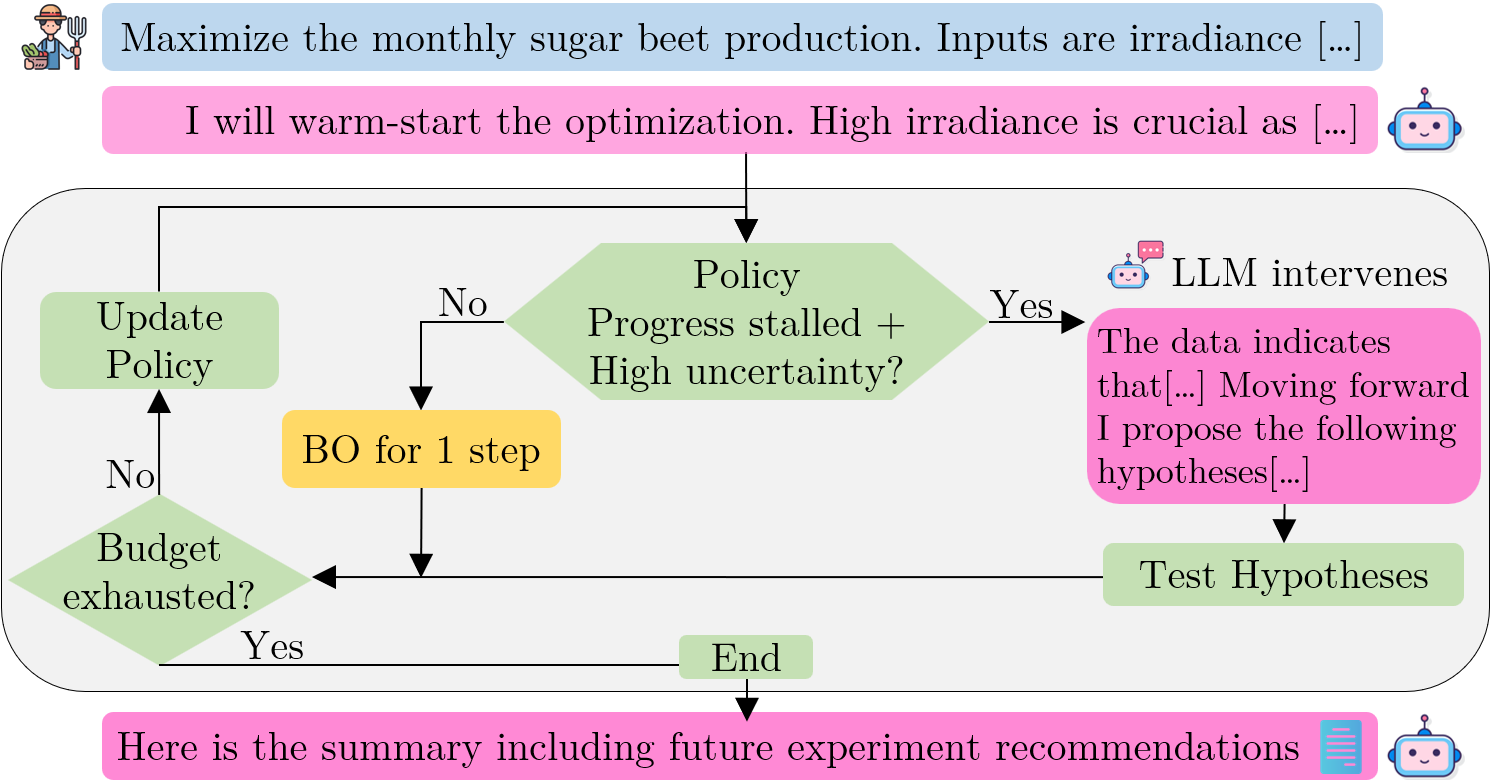}
    \caption{The BORA framework. Icons from~\protect\cite{c:flaticon_icons}.}
    \label{fig:bora_illustration}
\end{figure}

Exploring large experimental design spaces requires intelligent navigation strategies because of the costly and time-consuming function evaluations involved. Bayesian optimization has been established as an optimal experimental design methodology across disciplines spanning chemistry~\cite{c:GRYFFIN}, solar energy production~\cite{c:Solar} and agronomy~\cite{c:BOAgronomy}. BO can be used to efficiently navigate combinatorially large landscapes, and to identify promising solutions in an active-learning setting. Typically, BO uses a probabilistic surrogate to approximate an expensive or unknown objective function $f$ while iteratively searching for a maximizer,
\begin{equation}
    x^* = \operatorname*{argmax}_{x \in \mathcal{X}} f(x),
    \label{BO}
\end{equation}
with $f:\mathcal{X} \to \mathbb{R}$ defined on the search domain $\mathcal{X} \subseteq \mathbb{R}^{d}$.

This surrogate -- often a Gaussian Process (GP)~\cite{c:GP} -- subsequently undergoes Bayesian updates as new data about the design space are acquired, according to a predefined acquisition policy or function $\alpha(\cdot)$, allowing for the refinement of predictions of the objective $f(x)$. In turn, this acquisition function suggests the next set of parameters for experiments with the highest expected utility, balancing exploration in new regions and exploitation in promising ones.

Despite its successful application to a plethora of scientific tasks, BO is frequently characterized by slow initial optimization phases due to random or Latin Hypercube~\cite{c:WarmStartBO} selection of initial samples. This can slow the search for combinatorially large spaces substantially. This highlights a fundamental challenge for standard BO; that is, the lack of inherent domain knowledge and contextual understanding of the problem at hand. Recently, BO variants have been proposed that are capable of injecting domain-specific knowledge into the search, either through structural problem characteristics~\cite{c:DKIBO} or by using human expert knowledge~\cite{c:HypBO}. The latter approach, sometimes known as `Human-in-the-Loop (HIL)', has drawn considerable recent attention, and it aims to infuse domain knowledge and human reasoning into BO workflows~\cite{c:CoExBO,c:BOKaski}. By leveraging expert insights in the form of hypotheses~\cite{c:HypBO}, preferences~\cite{c:preferenceBO} or specified priors over possible optima~\cite{c:piBO,c:Incorporating_Expert_Prior_Knowledge}, it is possible to enrich the optimization process and to direct searches to fruitful regions much faster. Thus, HIL methods have shown increased effectiveness and efficiency compared with data-only approaches.  In particular, hypothesis-based methods have shown gains in both performance and cost. Nonetheless, these HIL approaches can be human-capital resource-intensive because they require regular human interventions. Moreover, it is easy, even for domain experts, to lose track of the state of the art in fast-moving research areas and to ignore certain promising regions of the search space~\cite{c:HIL}.

To address these challenges, we propose the use of Large Language Models (LLMs)~\cite{c:TowardsLLMs} as a facilitating framework in black-box optimization to enrich searches with domain knowledge. Specifically, we have coupled an LLM with standard BO in a hybrid optimization framework that automatically monitors and regulates the amount of domain knowledge needed when the search becomes `trapped' in local minima. The algorithm capitalizes on the LLM's inherent in-context learning (ICL) capacity and reasoning mechanism to suggest, in the form of hypotheses, promising areas in the search space from which to sample. LLMs have been employed recently to address limitations in core BO methodologies, as well as HIL variants~\cite{c:LLAMBO,c:ADOLLM}. LLMs have the capacity to encode vast amounts of domain-specific and general knowledge and have demonstrated the ability to reason about relatively complex tasks through in-context learning~\cite{c:InContextICL,c:ICL} as well as in multidisciplinary domains such as chemistry~\cite{c:LLMsInChemistry}. However, due to their numerically agnostic design, LLMs lag behind traditional BO methods in systematically balancing exploration versus exploitation, and have proved unreliable in many practical scenarios~\cite{c:LLMPotential}. Recent attempts have been made to integrate LLMs with BO frameworks~\cite{c:SLLMBO,c:ADOLLM} but thus far, these have been limited to small problem sizes, such as hyper-parameter optimization, or situations where the optimal solution is proximal to a special value~\cite{c:LLMPotential}. Also, LLM/BO hybrids could be prohibitively costly for more complex queries, particularly if the LLM is deployed for every iteration in the optimization.

Here, we propose a language-based Bayesian Optimization Research Assistant, BORA, that enriches BO with domain knowledge and contextual understanding across a range of scientific tasks. We frame BORA as a hybrid framework that augments surrogate-based optimizers with uncertainty estimates by localizing areas of interest in the search space, guided by a knowledge-enriched LLM (Figure~\ref{fig:bora_illustration}). A heuristic policy regulates the LLM involvement in the optimization process, adaptively balancing rigorous stochastic inference with LLM-generated insights within a feasible budget of LLM computation and API usage limits. During the intervention stage, the LLM uses its domain knowledge and reasoning capabilities to comment on the optimization progress thus far, highlighting patterns observed and forming hypotheses that may yield more rewarding solutions. It then tests these hypotheses by proposing new samples that maximize the target objective. BORA is also designed to provide an effective user-optimizer interaction through its dynamic commentary on the optimization process. This promotes deeper insights from the user and, in the future, the option to intervene; for example, by either reinforcing or overriding certain insights from the LLM. To our knowledge, this is the first time that a rigorous, dynamic synergy of black-box BO with LLMs has been proposed in this context. We evaluated BORA on various synthetic functions and a pétanque gaming model, as well as real scientific tasks in chemical materials design, solar energy production, and crop production. BORA demonstrated significant improvements in search exploration, convergence speed, and optimization awareness. Compared to earlier techniques, our method shows significant efficiency gains and generalization beyond hyper-parameter optimization, emphasizing its potential for tackling real-world tasks.

The remainder of this paper is organized as follows. Section~\ref{sec:related_works} presents recent works about domain knowledge and LLM integration in BO and Section~\ref{sec:methodology} details our proposed methodology. Section~\ref{sec:experiments} analyzes and compares the performance of our algorithm against state-of-the-art methods across diverse datasets, with Section~\ref{sec:conclusions} concluding our work and discussing future directions.

\section{Related Works}
\label{sec:related_works}
To cope with non-convex optimization landscapes in science tasks, intelligent approaches have been proposed that focus on promising regions through adaptive exploration-exploitation strategies~\cite{c:HybridAckBO}, or `smooth out' the optimization landscape by enriching it with domain knowledge~\cite{c:Incorporating_Expert_Prior_in_BO_via_Space_Warping}. Notable examples include local BO methods that restrict the search space, such as ZoMBI~\cite{c:ZoMBI}, which aims to improve efficiency by focusing on local regions assumed to contain the optimum. Similarly, TuRBO~\cite{c:TuRBO} uses multiple independent GP surrogate models within identified trust regions and a multi-armed bandit strategy~\cite{c:MAB} to decide which local optimizations to continue. These approaches are well-suited to handling high-dimensional problems, but their potential is perhaps more limited in small budgets and highly multimodal spaces due to a lack of built-in domain knowledge.

Incorporating domain knowledge into BO can improve both its efficiency and its performance~\cite{c:CoExBO,c:GRYFFIN}. DKIBO~\cite{c:DKIBO} enhances BO's acquisition function with structural knowledge from an additional deterministic surrogate model to enrich the GP's approximation power. Others, such as ColaBO~\cite{c:ColaBO} and HypBO~\cite{c:HypBO}, allow users to inject their beliefs at the start to guide the optimization process. However, those methods keep the users' beliefs static and cannot refine them as the optimization progresses, even if they are wrong. Meanwhile, other HIL methods rely on frequent user inputs~\cite{c:savage2023} and for robotic experiments~\cite{c:MobileRoboticChemist}, for example, that run 24/7 in a closed-loop way, waiting for this human user input might become the rate-limiting step.

Recently, some studies have explored LLMs as standalone replacements for traditional optimizers due to their exceptional ability to solve complex problems in various domains~\cite{c:LLMChemist,c:LICO}. Methods like LLAMBO~\cite{c:LLAMBO} and OPRO~\cite{c:OPRO} use the generative and ICL capabilities of LLMs to propose solutions to optimization problems directly. LLAMBO mimics BO's structure and replaces its key components with LLMs. In OPRO, the LLM is iteratively prompted with the gathered optimization data as input and tasked to generate new solutions as output that are then evaluated. These methods are innovative but have focused so far on low-dimensional hyperparameter tuning and are not yet obviously suitable as a general framework for optimization tasks. Querying LLMs at all iterations also incurs a larger computational and financial footprint than traditional BO algorithms, particularly if reasoning models are used. Standalone LLM optimizers also lack the mathematical guarantees offered by traditional optimizers such as BO. 

In response to the limitations of using LLMs as standalone optimizers, hybrid approaches such as BoChemian \cite{c:BoChemian} have emerged that combine the strengths of LLMs to featurize traditional optimization methods. SLLMBO~\cite{c:SLLMBO} integrates the strengths of LLMs in warm-starting optimization, and it loops between LLM-based parameter exploitation and Tree-structured Parzen Estimator (TPE)’s exploration capabilities to achieve a balanced exploration-exploitation trade-off. This reduces API costs and mitigates premature early stoppings for more effective parameter searches. However, SLLMBO, like LLaMEA-HPO~\cite{c:LLaMEA-HPO}, is limited to hyperparameter tuning. Moreover, its LLM exploration / TPE exploitation cycle lacks dynamic adjustment because it is an alternating process fixed at a 50:50 balance. Another limitation is the risk of premature optimization termination in complex search spaces due to a strict early stopping mechanism. 

Our approach, BORA, shares similarities with the above studies by incorporating domain knowledge and adapting search mechanisms. However, BORA is distinguished by leveraging LLMs when they are most required, for online hypothesis generation \textit{and} for real-time commentary on optimization progress. Unlike static methods such as HypBO, which assume fixed human-injected soft constraints, our method refines the optimization trajectory based on the contextual insights given by the LLM. Moreover, BORA extends beyond previous hybrid approaches such as SLLMBO by introducing adaptive heuristics that intelligently modulate LLM involvement with BO to maximize optimization performance.

\section{Methodology}

\label{sec:methodology}
The BORA optimization framework is illustrated in Figure~\ref{fig:bora_illustration}. It is an automated hybrid BO-LLM synergy operating under a common GP whose parameters are updated as new points are sampled, either from BO or the LLM. A user-provided experiment description is used to initially contextualize the LLM which then warm-starts the optimization with proposed samples through its ICL capabilities. The optimization progresses by alternating BO and LLM runs that are accordingly triggered by performance plateaus. Our proposed framework employs an adaptive heuristic policy to (a) assess the need to invoke the LLM, (b) determine the type of LLM intervention needed, and (c) update the frequency of those interventions as the optimization progresses. We use Reinforcement Learning (RL) terminology in the manuscript to describe our approach, but we use hand-crafted policy update rules because learning generalized rules in the traditional sense~\cite{c:RLABO,c:MetaBO} would be impractical in Bayesian scientific optimization settings~\cite{c:CostBO}, which is the focus of this work. In its interventions, the LLM provides user interpretability via real-time commentary of the optimization progress and generates hypotheses to maximize the objective.

\subsection{LLM Comments and Hypotheses}
\label{s:prompt_engineering}
\begin{figure}[t]
    \centering
    \includegraphics[width=\columnwidth]{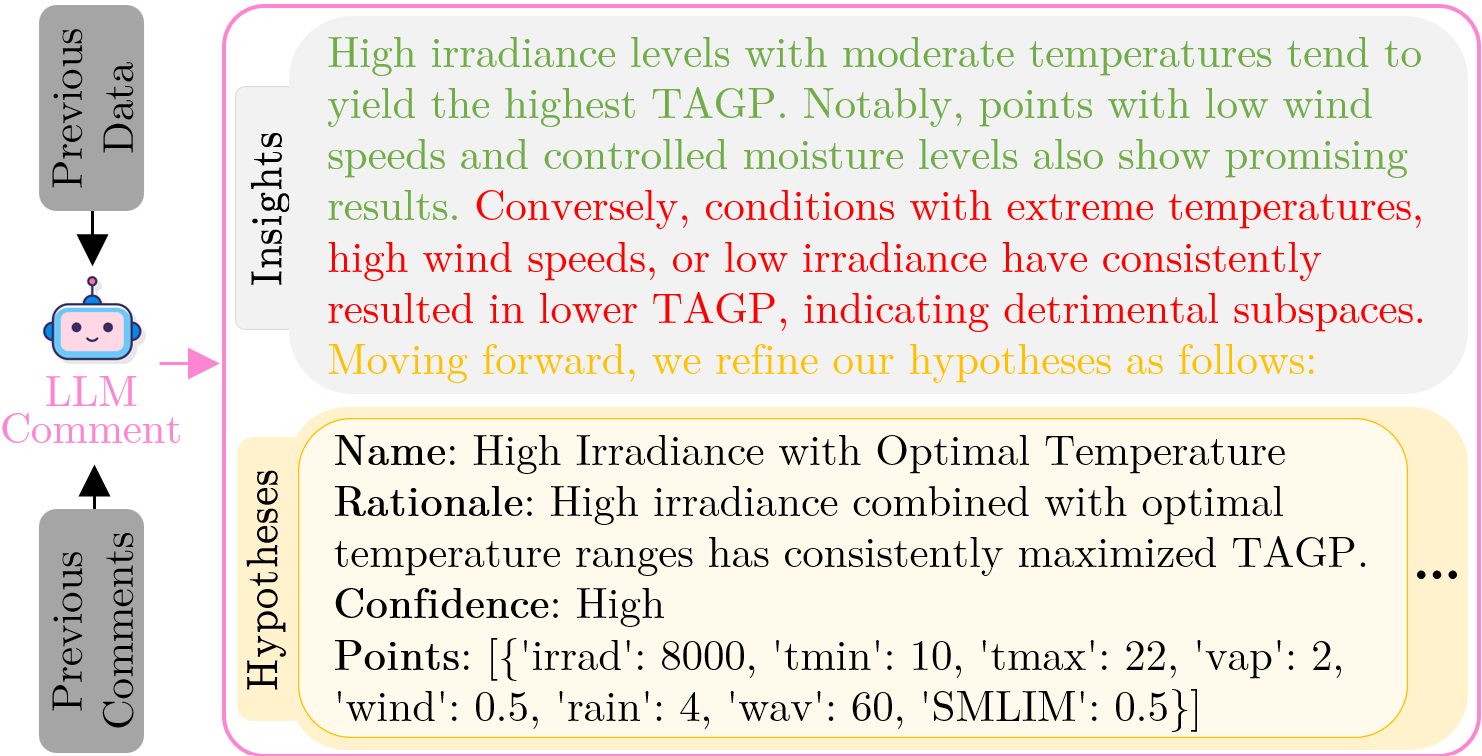}
    \caption{The LLM agent commenting and refining its hypotheses on the Sugar Beet Production experiment (complete comment in the SM). This experiment is detailed in Section~\ref{sec:experimental_setup}.}
    \label{fig:bora_comment}
\end{figure}

The LLM is prompt-engineered to return structured JSON responses that we call \emph{Comments} (for formatting details we refer the reader to the SM). The Comment object, illustrated in Figure~\ref{fig:bora_comment}, contains \emph{insights} into the optimization progress and potential reasons for stagnation, as well as a list of \emph{hypotheses} to remedy that stagnation. Each hypothesis includes a meaningful name, a rationale, a confidence level, and the corresponding input point to test it. Unlike in HypBO~\cite{c:HypBO} where hypotheses are defined as rather static regions of interest,
BORA dynamically builds hypotheses during the optimization process typically in the form of single search points through the LLM's ICL model. As demonstrated in LLAMBO~\cite{c:LLAMBO}, LLMs tasked with regression inherently perform an implicit ICL modeling of the objective function, estimating the conditional distribution $p(y|x;\mathcal{D})$, where $y$ is the target value at $x$. BORA extends this modeling by integrating all previously gathered data $\mathcal{D}$ and all the LLM's comments $\mathcal{C}$, enhancing the LLM surrogate to model $p(y|x;\mathcal{D}; \mathcal{C})$. From this augmented model, the LLM proposes hypotheses, exploring regions likely to improve on the current best observation $y_{\text{max}}$ and derived from the conditional probability $x^{\text{LLM}} \sim p(x|y>y_{\text{max}};\mathcal{D}; \mathcal{C})$.

\subsection{LLM Initialization}
\subsubsection{User-Provided Experiment Card}
To inform the LLM initially, the user prepares a comprehensive problem description following a standardized template that we refer to as the \emph{Experiment Card}. This card includes any details or context about the black-box function $f$ to be optimized, descriptions of its input variables, and the target variable to be maximized, along with any constraints that must be satisfied within the search space.
From the experiment card, the LLM is prompted to generate $n_\text{init}$ initial hypotheses for maximizing the target. This translates into $n_\text{init}$ initial points that are evaluated to form the initial dataset $\mathcal{D}_0 = \bigl\{\bigl( x_i, y_i = f(x_i)\bigr)\bigr\}_{i=1}^{n_\text{init}}$.

\subsection{Actions}
\label{s:actions}
BORA leverages an adaptive heuristic policy detailed in Section~\ref{s:policy} to choose one action from a set of three possible actions defined in the following paragraphs. The chosen action suggests at least one next point for evaluation, which is evaluated and added to the dataset. While the Vanilla BO action $a_{1}$ appends one sample $(x,y)$ to $\mathcal{D}_{t-1}$ at each step or iteration $t$, the LLM actions $a_2$ and $a_{3}$ add $n_{\text{LLM}} ~\text{and}~ n_{\text{LBO}}\ge 1$ samples, respectively. Hereon, we distinguish between $t$, the step number, and $i$, the sample index at step $t$, and denote with $\mathcal{S}_{t}=\{ x_{t}^{(i)} \}_{i=1}^{k}$ the set of $k$ points suggested by an action $a$ at step $t$.

\subsubsection{\texorpdfstring{\bm{$a_{1}$}}{a1} Continue with Vanilla BO}
The acquisition function is maximized to get the next promising point $\mathcal{S}_{t} = \{x_{t}^{(1)} \}$, which is then evaluated and added to the dataset to form $\mathcal{D}_{t} = \mathcal{D}_{t-1} \cup \{ (x, f(x))\ |\ x \in \mathcal{S}_{t}\}$.

\subsubsection{\texorpdfstring{\bm{$a_{2}$}}{a2} LLM Comments and Suggests \texorpdfstring{\bm{$n_{\text{LLM}}$}}{nLLM} Points}
A prompt containing the gathered data up to $\mathcal{D}_{t-1}$ and any previous comments $\mathcal{C}$ is given to the LLM. The LLM is then tasked to comment on the optimization progress in light of the new data and to update any previous hypotheses. The returned Comment contains the next points $\mathcal{S}_{t} = \{x_{t}^{(i)} \}_{i=1}^{n_{\text{LLM}}} \sim p(x|y>y_{\text{max}};\mathcal{D}_{t-1}; \mathcal{C})$ chosen by the LLM, which are then evaluated and added to the dataset to form $\mathcal{D}_{t}$.

\subsubsection{\texorpdfstring{\bm{$a_{3}$}}{a3} LLM Comments and Selects \texorpdfstring{\bm{$n_{\text{LBO}}$}}{nLBO} BO Points}
$a_{3}$ is a non-myopic, ICL step that focuses the LLM's attention on a high-quality set of $n_{\text{BO}}$ candidate points $\{x_{\text{BO}}^{(j)}\}_{j=1}^{n_{\text{BO}}}$, which are generated by maximizing the acquisition function. A prompt containing $\mathcal{D}_{t-1}$, $\mathcal{C}$, and $\{x_{\text{BO}}^{(j)}\}_{j=1}^{n_{\text{BO}}}$ are given to the LLM, which is then tasked to comment on the optimization progress in light of the new data and constrained to select the $n_{\text{LBO}}$ most promising BO candidates that best align with its hypotheses for maximizing the target objective. Setting $n_{\text{BO}}=5$ and $n_{\text{LBO}}=2$ empirically showed to offer enough diversity of hypothesized optima locations and ensure competitive performance overall. The returned Comment holds the $n_{\text{LBO}}$ selected points $\mathcal{S}_{t} = \bigl\{x_{t}^{(i)} \bigr\}_{i=1}^{n_{\text{LBO}}} \sim p(x \in \{x_{\text{BO}}^{(j)}\}_{j=1}^{n_{\text{BO}}} |y>y_{\text{max}};\mathcal{D}_{t-1}; \mathcal{C})$ that are then evaluated and added to the dataset.

\subsection{Adaptive Heuristic Policy}
\label{s:policy}
\subsubsection{Action Selection}
 BORA's policy helps it to make informed choices about engaging the LLM without relying on data-hungry RL algorithms, thus maintaining BORA's practicality and effectiveness in real-life scenarios. The optimization starts with the Vanilla BO action $a_1$, and the subsequent action selection depends on (a) the average uncertainty $\sigma_{\text{mean}}^{\text{GP}}$ of the common GP over the search space $\mathcal{X}$ to determine the necessity and type of LLM intervention, and (b) the BO performance plateau detection, as well as the performance success (or trust in) of the previous LLM interventions. When the GP's uncertainty is high and above a pre-defined threshold ($\sigma_{t, \text{mean}}^{\text{GP}} > \sigma_{t, \text{upper}}$), BO needs significant guidance from the LLM, triggering a complete `take-over' by the LLM in the search, suggesting new points informed by its own internal reasoning mechanism. As the GP's uncertainty decreases ($\sigma_{t, \text{lower}} \leq \sigma_{t, \text{mean}}^{\text{GP}} \leq \sigma_{t, \text{upper}}$), the LLM becomes less involved by relying only on BO suggested points, but still using its ICL capacity based on both $\mathcal{D}$ and $\mathcal{C}$ to select the most promising ones. When the GP's uncertainty is low enough ($\sigma_{t, \text{mean}}^{\text{GP}} < \sigma_{t, \text{lower}}$), BO has a better approximation of objective function's landscape and no longer needs guidance from the LLM. The rationale behind remark (b) is that the LLM should gain more trust as the LLM suggestions exhibit better performance, which in turn triggers the plateau duration to be re-defined as shorter, allowing the LLM to intervene more frequently. Conversely, if the LLM's so-far observed performance declines, its trust in itself diminishes and, consequently, its interventions are reduced, which results in longer plateau duration adjustments before invoking its assistance. In short, the action selection at each step $t$ follows the policy $\pi$ described below, where the GP parameters are updated after every action accordingly:
\begin{itemize}
    \item \underline{If} $\sigma_{t, \text{mean}}^{\text{GP}} < \sigma_{t, \text{lower}} $ or `no plateau' $\to$ \textbf{action $a_1$},
    \item \underline{Else if} $\sigma_{t, \text{mean}}^{\text{GP}} > \sigma_{t, \text{upper}}$ $\to$ \textbf{action $a_2$},
    \item \underline{Else} $\to$ \textbf{action $a_3$}.
\end{itemize}

\subsubsection{Selection Mechanism}
\paragraph{Uncertainties}
The above action selection is realized by calculating and updating in every step the uncertainties from a set of fixed $q$ monitoring points $x_{\text{mon}}^{(i)}$ that are randomly sampled before the optimization starts. Specifically,
\begin{align}
    \label{eq:uncertainties}
    \sigma_{t, \text{mean}}^{\text{GP}} =  \frac{1}{q} \sum_{i=1}^q \sigma_{t}\bigl(x_{\text{mon}}^{(i)}\bigr), \\
    \sigma_{t, \text{max}}^{\text{GP}} =\max\bigl(\sigma_{t-1, \text{max}}^{\text{GP}}, \max_{1 \leq i \leq q} \sigma_{t}(x_{\text{mon}}^{(i)})\bigr), \\
     \sigma_{t,\text{upper}} = 0.5 \times\sigma_{t, \text{max}}^{\text{GP}} ~\text{and}~ \sigma_{t,\text{lower}} = 0.3 \times\sigma_{t, \text{max}}^{\text{GP}},
\end{align}
where $\sigma_t(\cdot)$ represents the uncertainty of the GP at a given point in iteration $t$. Here, the 50\% and 30\% fractions serve as empirically tuned bounds that consistently balance BO exploitation with LLM exploration across diverse tasks.

\paragraph{Plateau Detection}
Another important part of the action selection mechanism in the proposed framework is the detection of performance plateauing in BO. A performance plateau is detected at step $t$ when
\begin{equation}
    \label{eq:plateau_detection}
    y_{j}^{\text{max}} < y_{j - 1}^{\text{max}} \times \bigl(1+\text{sign}(y_{j - 1}^{\text{max}})\times \gamma\bigr), \text{for all } j \in [t-m+1, t],
\end{equation}
where $y_{t}^{\text{max}}=\max\bigl(\{y|(x,y) \in \mathcal{D}_{t}\}\bigr)$. That is, if for the past $m$ consecutive BO steps, there is not enough performance improvement (w.r.t a set percentage $\gamma)$, then the LLM involvement is triggered. The plateau duration $m$ is initialized at $m_{\text{init}} =\lceil 2\sqrt{d} \rceil$, set to vary between $m_{\text{min}}=0$ and $ m_{\text{max}} =3 m_{\text{init}} $, and is automatically adjusted at every LLM intervention step $l$ (here $l$ counts the number of times actions $a_{2}$ or $a_{3}$ are invoked). The adjustment depends on the current `trust' $T_{l} \in [0, 1]$ BORA has on the LLM, which in turn relies on the LLM performance observed so far. Specifically
\begin{align}
    m_\text{adjustment} = \left\lfloor (T_{l} - T_{l-1}) \times \Delta_{\max} \right\rfloor, \\
    m \gets \text{clip}\left( m - m_\text{adjustment},\ m_{\text{min}},\ m_{\text{max}} \right),
\end{align}
where $\Delta_{\max}$ is the maximum allowed adjustment per step, here set to 15, and $\text{clip}(x, a, b)$ is a function that restricts $x$ to be within the bounds $[a, b]$. 

\paragraph{Trust Mechanism}
As noted above, the plateau adjustment relies on an adaptive trust mechanism that regulates the trust in the LLM as defined by a `trust score' calculated on previous performances. That is, at each step $t$ where the LLM suggests (or selects) $\mathcal{S}_{t}=\bigl\{x_{t}^{(i)}\bigr\}_{i=1}^{k}$, the trust score is updated based on the following reward function
\begin{equation}
    \label{eq:improvement}
    r_{l} = \max\bigl(\{f(x)|x \in \mathcal{S}_{t}\}\bigr) - y_{t-1}^{\text{max}}
\end{equation}
First, an intervention score, ranging in $[0, 1]$, reflects the utility of those LLM suggestions in finding a new optimum with respect to the reward function in Eq.~\eqref{eq:improvement} as
\begin{align}
    \text{score}_{\text{interv}}^{(l)} &= 
    \begin{cases} 
    1, & \text{if } r_{l} > 0, \\
    \frac{1}{1 + e^{-\frac{r_{l}}{|y_{t-1}^{\text{max}}| + \epsilon}}}, & \text{if } r_{l} \leq 0,
    \end{cases} \label{eq:score}
\end{align}
where $\epsilon=10^{-6}$ is a small constant to handle cases where 
$y_{t-1}^{\text{max}} = 0$. By normalizing $r_{l}$ with $|y_{t-1}^{\text{max}}|$, the trust score becomes more sensitive to relative changes rather than absolute changes. This is particularly useful in domains where the magnitude of $y$ varies widely, making it robust across scales. Then, this intervention score is added to $\mathcal{H} \gets \mathcal{H} \cup \{\text{score}_{\text{interv}}^{(l)}\}$, keeping track of the previous intervention scores. Note that to reflect an initially optimistic view of the LLM, $\mathcal{H}$ is initialized as $\{0.9\}$, i.e., $\text{score}_{\text{interv}}^{(0)} = 0.9$. Finally, an average rolling trust score $T_{l}$ is subsequently calculated as the average of the intervention scores in $\mathcal{H}$ over a sliding window $W$ of the three most recent intervention scores as
\begin{equation}
    \label{eq:trust_score}
    T_{l} = \frac{1}{W} \sum_{i=|\mathcal{H}| - W}^{l} \text{score}_{\text{interv}}^{(i)} \text{ where }W=\min(|\mathcal{H}|,3).
\end{equation}
The complete BORA framework is described in Algorithm~\ref{alg:BORA}. Details about the LLM prompt engineering, reflection strategies, and fallback mechanisms can be found in the SM.

\begin{algorithm}[tb]
    \caption{BORA}
    \label{alg:BORA}
    \textbf{Input}: Experiment card, Number of initial samples $n_\text{init}$, Maximum number of samples $i_{\text{max}}$
    \textbf{Output}: {$y_{\text{max}}$}, LLM comments $\mathcal{C}$ and final report
    
    \begin{algorithmic}[1]
    \STATE LLM generates initial samples $\mathcal{D}_0 \gets \{(x_{i}, f(x_i))\}_{i=1}^{n_\text{init}}$;
    \STATE Initialize the GP surrogate model with $\mathcal{D}_0$;
    \STATE Initialize policy parameters $\sigma_{0, \text{mean}}^{\text{GP}}$, $\sigma_{0, \text{max}}^{\text{GP}}$, $\sigma_{0, \text{upper}}$, $\sigma_{0, \text{lower}}$, $m$, $\mathcal{H}= \{0.9\}$, $\gamma=0.05$, $n_{\text{BO}}=5$, $n_{\text{LBO}}=2$;
    \STATE Initialize sample index $i = n_\text{init}$, step $t = 1$, $\mathcal{C}=\{\}$;
    \WHILE{$ i <  n_\text{init} + i_{\text{max}}$}
        \IF {$\sigma_{t, \text{mean}}^{\text{GP}} < \sigma_{t, \text{lower}}$ or `no plateau'}
        \STATE $a = a_{1}$,~$\mathcal{S}_{t} = \{x_{t}^{(1)}\} $;
        \ELSIF {$\sigma_{t, \text{mean}}^{\text{GP}} > \sigma_{t, \text{upper}}$ }
        \STATE $a = a_{2}$,~$\mathcal{S}_{t} = \{x_{t}^{(k)}\}_{k=1}^{\min(n_{\text{LLM}},i_{\text{max}} + n_\text{init} - i)}$;
        \ELSE
        \STATE $a = a_{3}$,~$\mathcal{S}_{t} = \{x_{t}^{(k)}\}_{k=1}^{\min(n_{\text{LBO}},i_{\text{max}} + n_\text{init} - i)}$;
        \ENDIF
        \STATE Update dataset $\mathcal{D}_{t} \gets \mathcal{D}_{t-1} \cup \{ (x, f(x))|x \in \mathcal{S}_{t} \}$;
        \STATE Update $y_{\text{max}}$ as the maximum $y$ value in $\mathcal{D}_{t}$;
        \STATE Update GP, policy parameters and trust mechanism;
        \STATE $t \gets t+1$ and $i \gets i+k$;
    \ENDWHILE
    \STATE LLM generates a final report.
    \end{algorithmic}
\end{algorithm}

\section{Experiments}
\label{sec:experiments}
We validated BORA's performance against current state-of-the-art methods on both synthetic functions and various real-world tasks, with dimensionality ranging from 4 to 15 independent variables. Section~\ref{sec:experimental_setup} outlines the experimental setup while Section~\ref{sec:results} highlights the results. Details on the benchmarks, the method implementations, and the reproducibility details can be found in the SM. The source code is available at https://github.com/Ablatif6c/bora-the-explorer.

\subsection{Experimental Setup}
\label{sec:experimental_setup}
\subsubsection{Synthetic Function Benchmarks}
\begin{figure}[tb]
    \centering
    \includegraphics[width=\columnwidth]{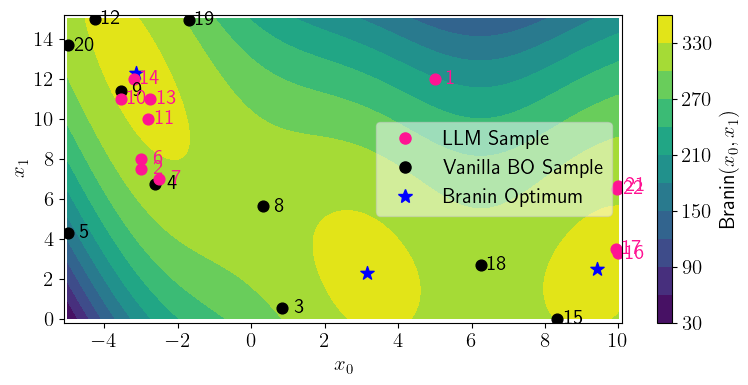}
    \caption{Visualization of BORA maximizing Branin 2D (which contains three global maxima) under a budget of 22 optimization steps (numbered points). Black samples were suggested by the BO action $a_{1}$, while pink ones came from the LLM actions $a_{2}$ and $a_{3}$.}
    \label{fig:bora_search}
\end{figure}

\begin{itemize}
    \item \textbf{Branin (2D)}: A function with a global maximum occurring in three distinct locations as shown in Figure~\ref{fig:bora_search}. The input space bounds are $x_{0} \in [-5, 10]$ and $x_{1} \in [0, 15]$. 
    \item \textbf{Levy (10D)}: A function with a highly rugged landscape. All inputs are bounded in $[-10, 10]$ with the maximum at $[1,\dots,1]$.
    \item \textbf{Ackley (15D)}: A challenging high dimensional function with several local maxima. Input bounds are $[-30, 20]$ with the maximum at  $[0,\dots,0]$.
\end{itemize}
Note that the names of these functions in the experiment card were anonymized to `mathematical function' to prevent the LLM from recognizing them by name.
\subsubsection{Real-World Application Benchmarks}
\begin{itemize}
    \item \textbf{Solar Energy Production (4D)}: Maximizing the daily energy output of a solar panel by optimizing panel tilt, azimuth, and system parameters~\cite{c:pvlib}.
    \item \textbf{Pétanque Game (7D)}: A ball is thrown to hit a target. The goal is to maximize the score, which is inversely proportional to the target distance miss, by tuning the throw position, angles, velocity, and spins.
    \item \textbf{Sugar Beet Production (8D)}: Maximizing the monthly sugar beet Total Above Ground Production (TAGP) in a greenhouse by tuning the irradiance, and other weather and soil conditions~\cite{c:PCSE}.
    \item \textbf{Hydrogen Production (10D)} Maximizing the hydrogen evolution rate (HER) for a multi-component catalyst mixture by tuning discrete chemical inputs under the constraint that the total volume of the chemicals must not exceed 5 mL. Note that due to the discrete and constrained nature of the problem, we adapted all compared methods accordingly to account for this, by employing the bespoke implementation of~\cite{c:MobileRoboticChemist}. Dataset acquired from~\cite{c:HypBO}.
\end{itemize}

\subsubsection{BORA}
We implemented BORA using OpenAI's most cost-effective model at the time, GPT-4o-mini~\cite{c:openai2025gpt4omini}, which was not fine-tuned in our effort to make BORA more accessible to users with limited resources. For the BO action implementation, the GP uses a Matérn kernel, and the acquisition function is EI. We set $q=5{,}000$ for $\sigma_{t,\text{mean}}^{\text{GP}}$.

\subsubsection{Baselines}
\begin{itemize}
    \item \textbf{Random Search}: Unbiased exploration baseline.
    \item \textbf{BayesOpt}~\cite{c:BayesOptGitHub}: Example of vanilla BO.
    \item \textbf{TuRBO}~\cite{c:TuRBO} with a single trust region.
    \item \textbf{ColaBO}~\cite{c:ColaBO} that uses a single static expert given-prior over the optimum to guide the
optimization process.
    \item \textbf{HypBO}~\cite{c:HypBO} that uses multiple static expert-given-promising regions to guide the optimization.
    \item \textbf{LAEA}~\cite{c:LAEA}, a hybrid LLM-Evolutionary Algorithm method.
\end{itemize}
For ColaBO and HypBO, to avoid the impracticality of relying on humans to provide inputs for multiple trials across all experiments, we used the LLM GPT-4o-mini to generate the `human' inputs. Likewise, we used the same task description prompts as used for BORA, to ensure consistency. For HypBO on the Hydrogen Production experiment, we employed the most realistic hypothesis used in~\cite{c:HypBO}, namely `What They Knew', which encapsulates any human knowledge available prior to the execution of the experiment.

\begin{figure}[tb]
    \centering
    \includegraphics[width=\columnwidth]{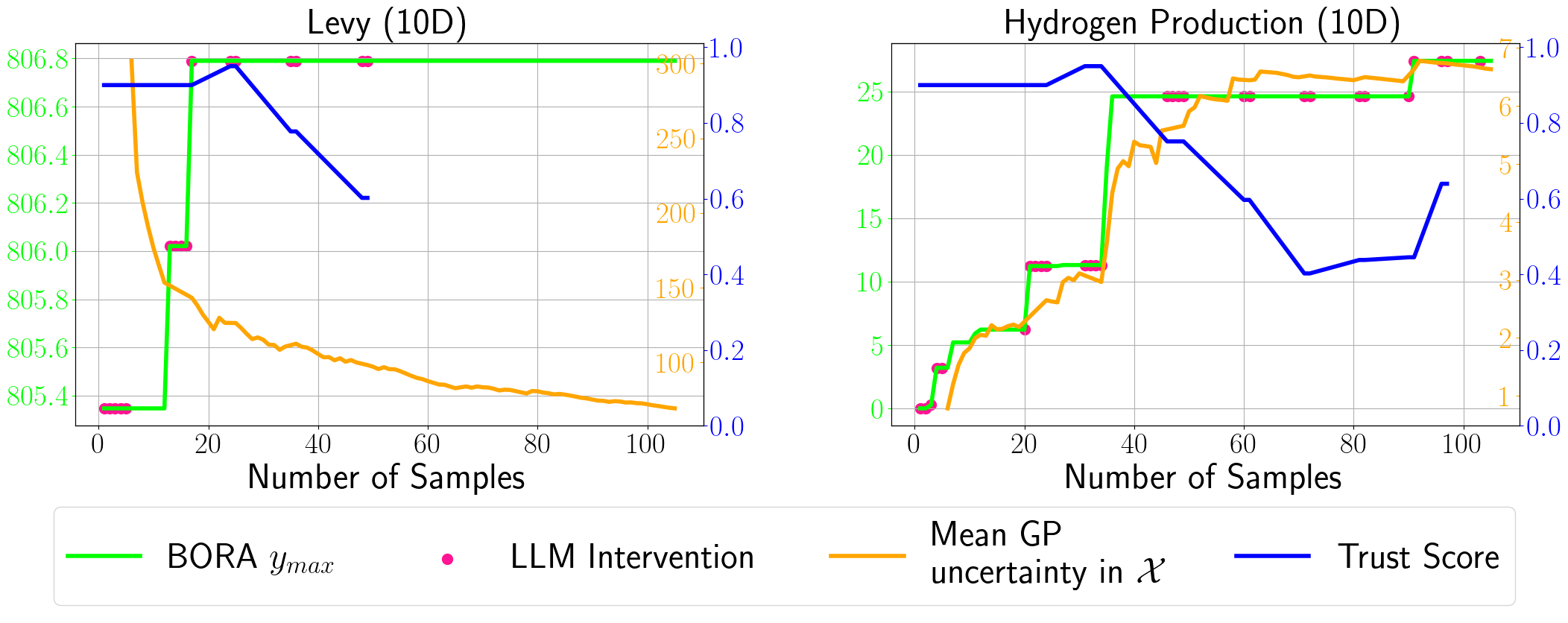}
    \caption{LLM intervention monitoring during a BORA run on 10D datasets Levy (left) and Hydrogen Production (right). Mean uncertainty and Trust scores are also overlayed to highlight their interrelationships.}
    \label{fig:llm_influence}
\end{figure}
\subsubsection{Experimental Protocol}
The optimization performance was measured using the maximum objective value found so far, the cumulative regret, and statistical tests to measure significance. The maximum number of samples was set to 105 to account for realistic budgets with expensive functions. Average results of 10 repeated trials with distinct random seeds are reported. All methods were initialized with $n_{\text{init}}=5$ initial samples apart from LAEA, for which we used 15 initial samples to keep the same number of evaluations to population size ratio as in~\cite{c:LAEA}.

\begin{figure*}[tb]
    \centering
    \includegraphics[width=\textwidth]{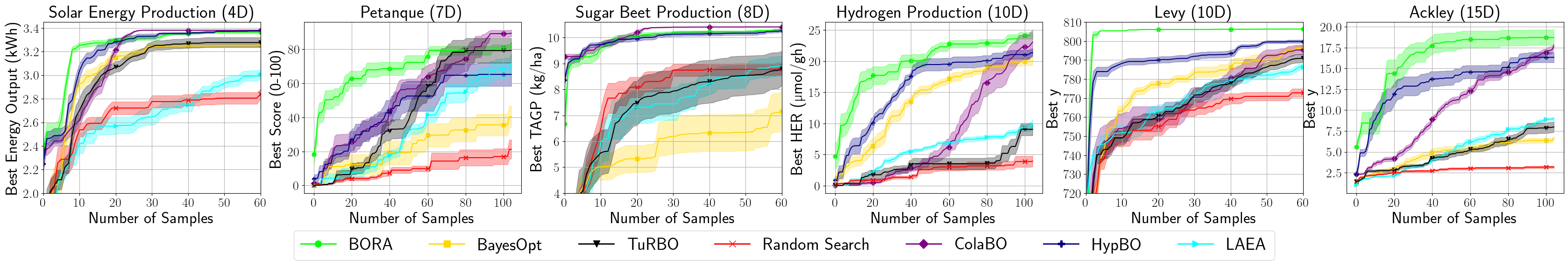}
    \caption{BORA vs Baselines on six experiments. Solid lines show average values while shaded areas indicate $\pm 0.25$ standard error. For visual clarity, some plots are zoomed in to show results up to 60 iterations, as the trends mostly stabilize beyond this point. Full results in SM.}
    \label{fig:all_experiments}
\end{figure*}

\begin{figure}[t]
    \centering
    \includegraphics[width=\columnwidth]{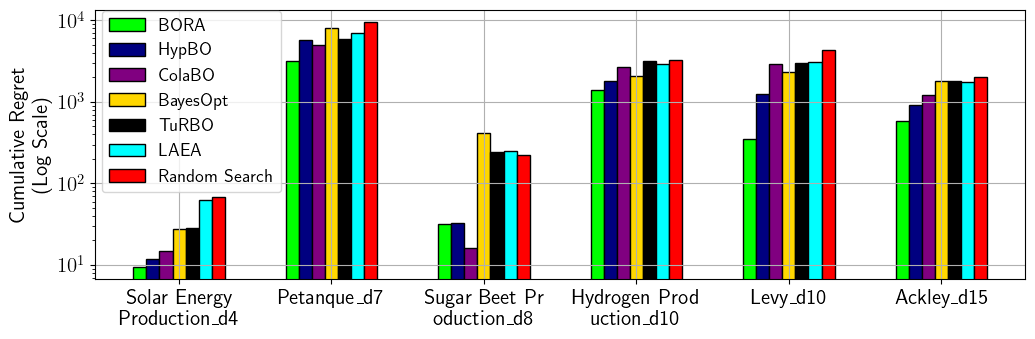}
    \caption{Comparison of BORA vs. Baselines' cumulative regrets in the six experiments.}
    \label{fig:cum_regrets}
\end{figure}

\subsection{Results}
\label{sec:results}

\subsubsection{Synthetic Functions}
\label{sec:synthetic_functions}
Figure~\ref{fig:bora_search} illustrates the exploration strategy of BORA on the Branin (2D) function. The LLM interventions helped to uncover two out of the three possible locations of the global maximum of Branin. This is further illustrated in Figure~\ref{fig:all_experiments}, which shows that BORA outperforms the baseline comparisons for the higher-dimension functions Levy (10D) and Ackley (15D). A key advantage of BORA is its LLM-informed initial sampling. For mathematical functions, BORA systematically suggests initializing at critical points such as the edges, central points, or other remarkable points like $[0,\dots,0]$. This strategy is particularly well-suited for the Levy function, whose search space bounds are symmetric, and its optimum is at $[1,\dots,1]$, almost always converging in its initialization stage. However, that strategy is less beneficial on the Ackley function because its bounds are asymmetric. Despite that, the LLM's ability to reflect and learn from the previous samples appears to help mitigate any unfavorable initializations. While HypBO demonstrates a similar benefit through its initial sampling in hypothesized regions, its performance is comparatively weaker because it relies on random initial sampling within these regions, resulting in less effective exploration of the search space. For ColaBO, which only works with a single input prior, the prior tended to be around one of the edges, which overall limits its convergence speed. Additionally, the left panel of Figure~\ref{fig:llm_influence} shows how BORA's iterative hypothesis generation, informed by previous data, helps mitigate stagnant optimization, and discards the LLM when it is no longer needed. Notably, a sharp drop in the GP uncertainty is evident when vanilla BO is used due to is proved exploration-exploitation guarantees, as opposed to the less rigorous LLM where the uncertainty is bound to its inherent sampling strategy. Nevertheless, the dynamic synergistic effect of BO coupled with updated LLM hypotheses allows for faster convergence overall, in comparison to other baselines as illustrated in the last two bar plots of Figure~\ref{fig:cum_regrets}.

\subsubsection{Real-World Applications}
BORA also exhibits superior performance across diverse real-world optimization problems, following the trends observed in the synthetic benchmarks. As shown in Figure~\ref{fig:all_experiments}, while BORA's initial sampling is on par with other input-based baselines for the Solar Energy and Sugar Beet Production experiments, its overall performance on all experiments surpasses the baselines significantly. This is particularly evident in the 7D Pétanque experiment, where BORA's diverse initial hypotheses based on trajectory dynamics led to a remarkable gain in score of 35 in the early stages compared to the baselines. This knowledge and context-based input bridges the knowledge gap typically encountered in early-stage optimization, providing BORA with a critical advantage, as shown in Figure~\ref{fig:cum_regrets}. A similar effect is also observed in the Hydrogen Production experiment as illustrated in the right panel of Figure~\ref{fig:llm_influence}. In addition, in the later stage of the optimization the LLM further pushes the optimization to uncover new optima after the progress had stalled, thus gaining more trust. The increasing trend in the GP uncertainty here is a side-effect of the continual interventions of the LLM, which translates to a rather explorative and less exploitative strategy based on its inherent domain reasoning around cumulatively accrued scientific data. This goes beyond some of the near-instant convergence noted in the optimization of the synthetic functions because they are proximal to a special value. Figure~\ref{fig:bora_comment} illustrates this by showing how the LLM reflects on the progress and generates hypotheses on the Sugar Beet Production experiment. While other baselines, particularly knowledge-based methods such as HypBO and local BO approaches such as TuRBO, demonstrate improved performance as the optimization progresses and more data is gathered, they often struggle to match BORA's sustained performance as the 105-sample budget mark is approached. In the Hydrogen Production experiment, this adaptive strategy ultimately achieved a 47\% reduction in cumulative regret compared to ColaBO, demonstrating BORA's faster convergence and robustness in navigating complex, high-dimensional search spaces. To assess the significance of the performance difference w.r.t mean cumulative regret between BORA and its best two competitors, we performed a sign test which revealed that BORA performs consistently better than HypBO with a p-value of 0.02 at a 95\% confidence level with a Bonferroni correction~\cite{c:bonferroni}, but not against ColaBO with a p-value of 0.20, yet still outscoring it in 5 out of 6 tasks. The superior performance of the hybrid approach in BORA was further validated by ablation studies that used only the LLM (action $a_{2}$) for optimizing Hydrogen Production (10D) and the Ackley function (15D) (see SM). While performing quite well in the initial stages for these two problems, the use of the LLM alone was ultimately less effective than the dynamic hybrid BO/LLM approach in BORA. We emphasize that these results do not mean that LLMs are `smarter' than domain experts. Rather, they highlight BORA's ability to update and refine its hypotheses based on new data, which is not possible in the HypBO implementation~\cite{c:HypBO}, while also fostering user engagement by generating real-time optimization progress commentary and a final summary report (see SM). One potential limitation of BORA, however, is the stochastic nature of the LLM reasoning, which can diverge considerably even with identical prompts.

\section{Conclusions}
\label{sec:conclusions}
This work introduces BORA, the first optimization framework to integrate BO with LLMs in a cost-effective dynamic way for scientific applications. BORA leverages the reasoning capabilities of LLMs to inject domain knowledge into the optimization process, warm-starting the optimization and enabling hypothesis-driven exploration and adaptive strategies to navigate complex, non-convex search spaces. It addresses key limitations in traditional BO methods, including slow initialization, local minimum entrapment, and the lack of contextual understanding. Notably, BORA outperformed BO with the addition of static expert-knowledge-derived hypotheses in a challenging 10D chemistry experiment, Hydrogen Production, highlighting its potential as a collaborative AI tool to support and enhance expert decision making.  Future directions will include refining BORA's meta-learning strategies using multi-agent LLMs and exploring its effectiveness in multi-objective, multi-fidelity optimization scenarios.

\section*{Acknowledgments}
The authors acknowledge financial support from the Leverhulme Trust via the Leverhulme Research Centre for Functional Materials Design. The authors also acknowledge the AI for Chemistry: AIchemy hub for funding (EPSRC grant EP/Y028775/1 and EP/Y028759/1). This project has received funding from the European Research Council under the European Union’s Horizon 2020 research and innovation program (grant agreement no. 856405). AIC thanks the Royal Society for a Research Professorship (RSRP\textbackslash S2\textbackslash 232003).

\bibliographystyle{named}
\bibliography{ijcai25}

\clearpage
\section*{Supplementary Material}
\appendix
\section{Extended Results and Ablations}
\subsubsection{Main results}
\begin{figure}[h]
    \centering
    \includegraphics[width=\columnwidth]{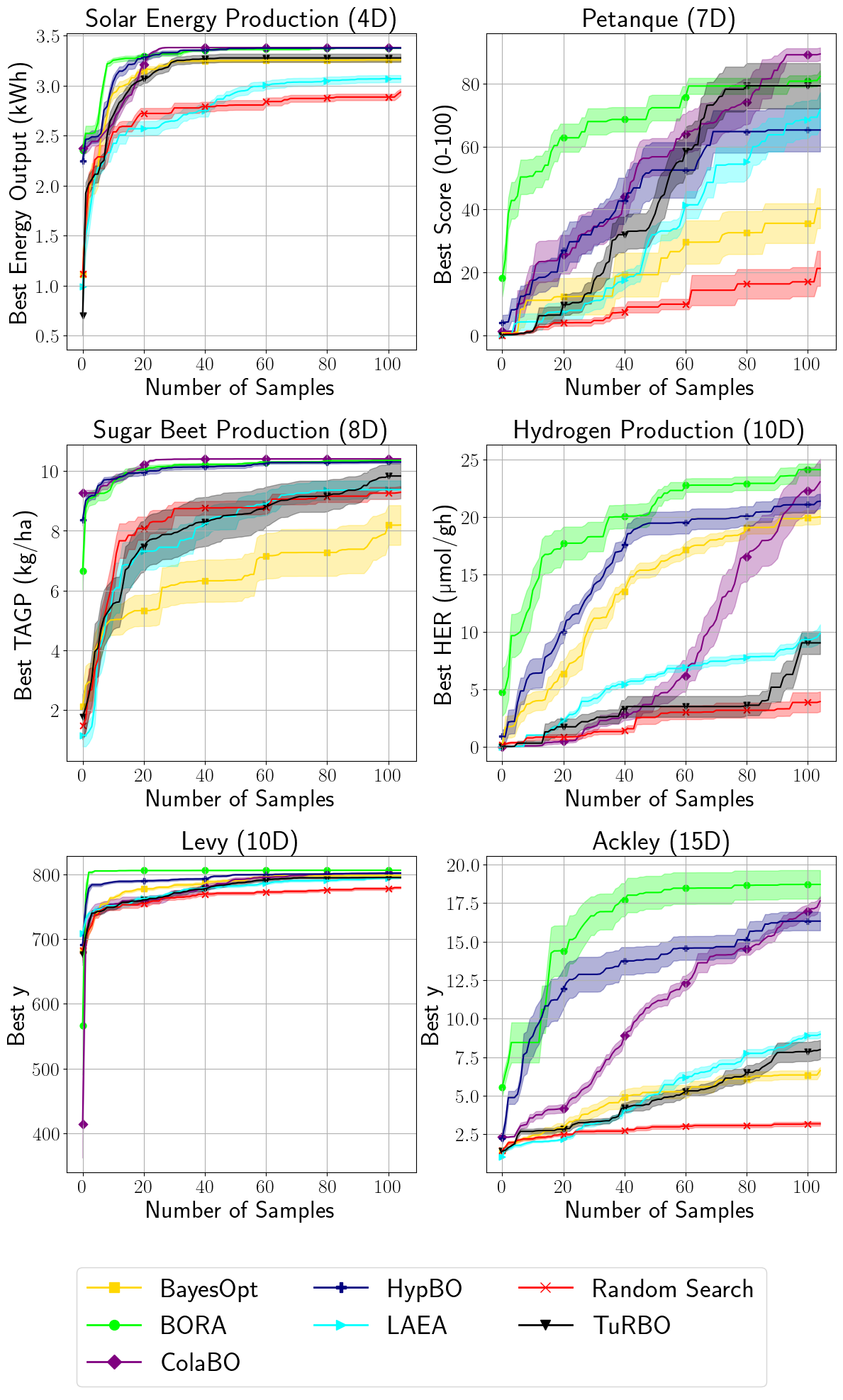}
    \caption{Full results of the BORA vs Baselines on the six experiments. Solid lines show average values while shaded areas indicate $\pm 0.25$ standard error.}
    \label{fig:full_main_results}
\end{figure}

Figure~\ref{fig:full_main_results} presents the complete, unzoomed plots (all 105 iterations) comparing BORA with the baselines showing the maximum objective value found over the six experiments.

\subsubsection{BORA vs. LLM-Only Optimizer}
To further evaluate BORA's contributions, we compared it with an LLM-only optimizer, which effectively isolates the LLM component of BORA and functions as an independent optimizer. This comparative analysis allowed us to determine the degree to which the incorporation of Bayesian Optimization (BO) principles enhanced performance beyond the capabilities of the LLM alone. LLM-Only uses identical prompts, with only minor modifications in the role-based prompting mechanism to ensure that it acknowledges its status as the sole optimizer without anticipating BO interventions. It employs the same hypothesis generation mechanism as BORA's LLM action $a_{2}$, relying exclusively on the outputs of the LLM for optimization. Additionally, LLM-Only uses the same reflection and fallback mechanisms. In the rare event where the LLM fails to produce a valid hypothesis, the corresponding hypothesis point is replaced with a randomly generated point to maintain continuity.

\paragraph{Experimental Setup}
We evaluated BORA and LLM-Only on the more complex synthetic function Ackley (15D) and on the most demanding highest-dimensional real-world task, Hydrogen Production (10D), and reported the maximum objective value found. To ensure comparability with the experimental protocol outlined in the manuscript, LLM-Only was executed with an optimization budget of 105 samples and repeated 10 times.

\begin{figure}[!b]
    \centering
    \includegraphics[width=\columnwidth]{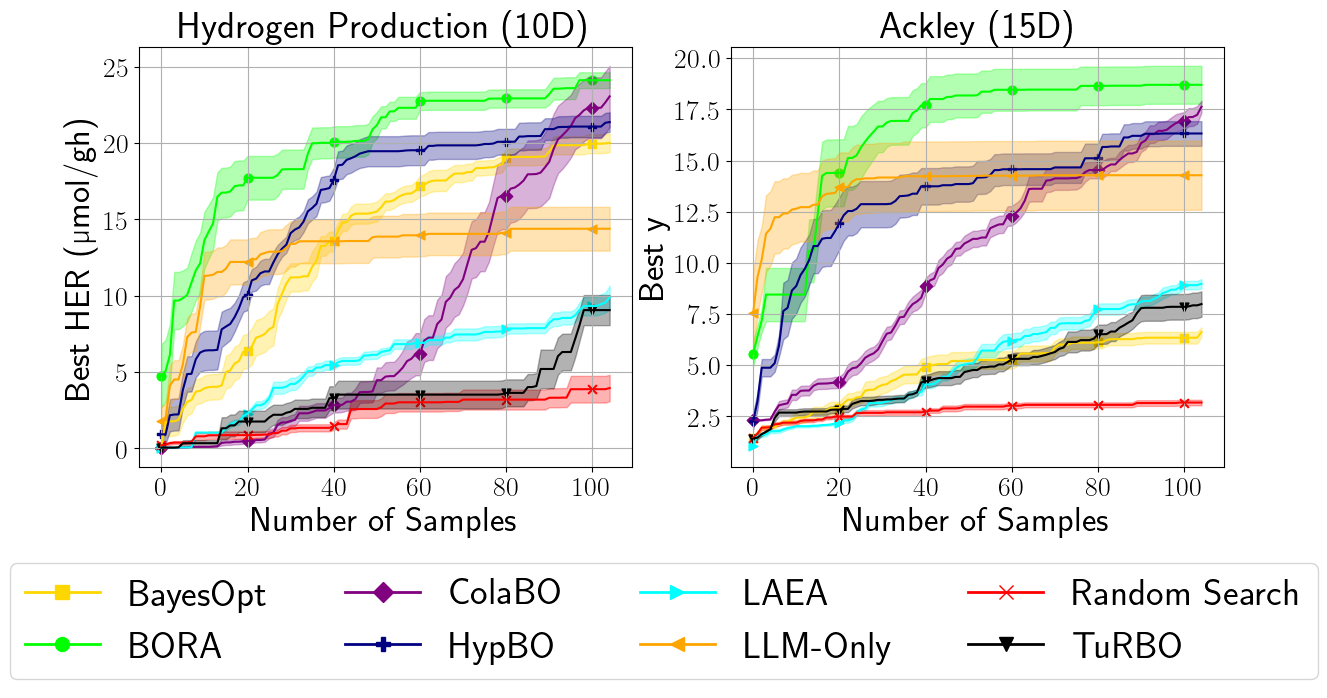}
    \caption{Comparison of BORA vs. LLM-Only on the Hydrogen Production (10D) and Ackley (15D) benchmarks.}
    \label{fig:bora_vs_llm}
\end{figure}

\paragraph{Results}
Figure~\ref{fig:bora_vs_llm} shows that LLM-Only exhibits competitive performance during the early stages of optimizing both Ackley (15D) and Hydrogen Production (10D). This achievement can be attributed to LLMs' ability to perform inductive biases, which are used during hypothesis generation. However, LLM-Only fails to sustain its initial performance and quickly stagnates, falling behind BORA significantly in the later stages of the optimization. This stagnation highlights the limitations of relying solely on LLM inference to optimize high-dimensional spaces. In particular, LLMs lack the balanced exploration-exploitation trade-offs that are inherent to BO-based methods, which are achieved through posterior uncertainty modeling. By contrast, BORA, as a hybrid BO/LLM approach, dynamically merges structured sampling from BO with the LLM's contextual understanding, facilitating sustained improvements as the optimization progresses. Notably, BORA achieved up to a 67\% improvement in Hydrogen Production and a 31\% improvement in the Ackley benchmark based on the maximum objective values discovered. While BORA considerably outperforms LLM-Only, it is interesting to note that while LLM-Only struggles to sustain its initial, early-stage performance, it still outperforms half of the baselines studied here in this small-budget setting.

\subsubsection{Ablation Studies}
In this section, we assess BORA's flexibility in dynamically switching between LLM-dominated search strategies and BO-dominated ones in the optimization. A key parameter that regulates this is the plateau window size $[m_{\text{min}}=0,\quad m_{\text{max}}=3 m_{\text{init}}]$, conditioned by the chosen initial value of the plateau $m_\text{init}$. In this ablation study, we test how different values of $m_\text{init}$ affect BORA's performance.

We select  $m_\text{init} = 2$, which translates into a narrower plateau window of $[0,\ 6]$ and thus increases the number of plateaus encountered during the optimization, leading to more frequent LLM interventions. Additionally, we test $m_\text{init} = 32$, which corresponds to fewer LLM interventions and more BO steps. We test BORA's performance on both of these scenarios on the Ackley (15D) function. Note that we also include in the comparison:
\begin{itemize}
    \item The default BORA setting of $m_{\text{init}} =\lceil 2\sqrt{d} \rceil$ where $d$ is the experiment dimensionality, corresponding to $m_{\text{init}} = 8$ for Ackley (15D).
    \item BayesOpt as the Vanilla BO baseline.
    \item LLM-Only as the pure LLM baseline.
\end{itemize}

\begin{figure}[!t]
    \centering
    \includegraphics[width=\columnwidth]{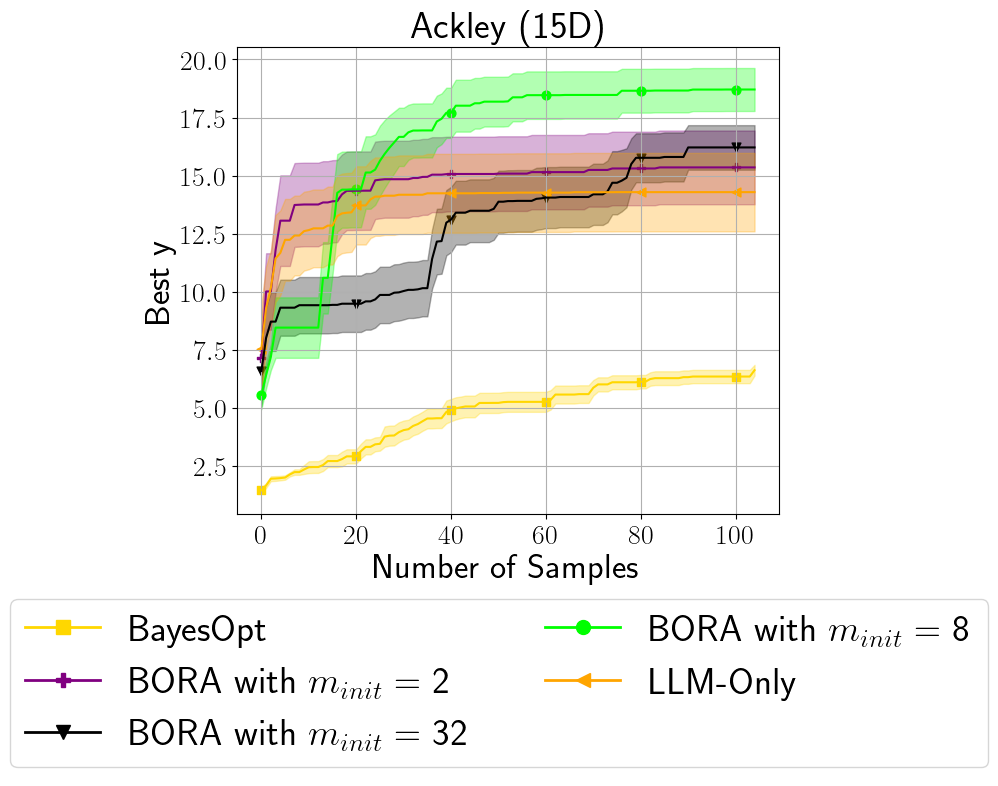}
    \caption{Comparison of BORA's performance for different values of plateau window size on the Ackley (15D) function for 10 trials.}
    \label{fig:bora_m_ablation_study}
\end{figure}

Figure~\ref{fig:bora_m_ablation_study} shows the results of the comparison. Out of 10 runs, we observe that the algorithm behaves closer to the Vanilla BO as the plateau window size increases, as one might expect. Likewise, BORA behaves closer to the LLM-only model as the window size is reduced. This highlights BORA's flexibility in adapting dynamically to different problem types that might require more or less input from the LLM.

\subsubsection{Cost of BORA Runs}
We report the total cost of running BORA and LAEA in the manuscript experiment section, both using GPT-4o-mini as the LLM component, which covers 6 benchmarks across 10 trials with a 105-sample budget. The total cost of running BORA amounts to about 5 US Dollars (USD) at today's prices, while the cost for LAEA with an initial population size of 15 is around 41 USD, considering the LLM costs only. This cost difference arises from the inherent design of the two methods. As an LLM-BO hybrid, BORA alternates between LLM-based and BO-based sampling via its adaptive policy. Notably, the LLM is more involved in optimization when it proves trustworthy. In contrast, LAEA uses the LLM as a surrogate model to evaluate the fitness of the individuals in the population. Given that at each iteration during the optimization, LAEA uses the LLM to estimate the fitness of each individual in the population, this translates into a much more frequent LLM call, leading to increased computational and inference costs. To put this into perspective for a population size of $\text{pop\_size} = 15$ individuals, and a maximum number of objective function evaluations of $n_{\text{evals}} = 105$, LAEA requires at least 
\[
n_{\text{LLM\_calls}}=(n_{\text{evals}}-\text{pop\_size}) \times \text{pop\_size}=1350,
\]
plus the LLM calls incurred by the reflection strategies. These findings demonstrate that BORA's strategic use of LLMs, beyond performing significantly better, has much smaller LLM inference and cost footprints, noting that the use of LLMs also has a significant (and growing) energy cost.

\begin{figure}[!b]
    \centering
    \includegraphics[width=\columnwidth]{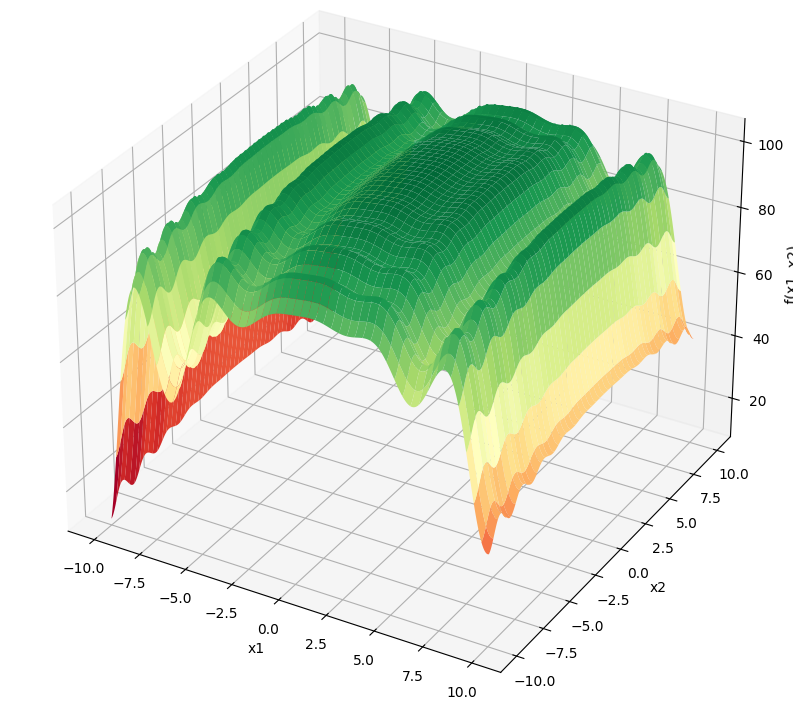}
    \caption{Landscape of 2D Levy Function}
    \label{fig:levy}
\end{figure}

\begin{table*}[t]
    \centering
    \begin{tabular}{lllr}
    \toprule
    Variable & Unit & Description & Bound \\
    \midrule
    Tilt Angle         & °              & Angle of the solar panel relative to the horizontal plane. & [0, 90] \\
    Azimuth Angle      & °              & Angle describing the direction the panel is facing on the horizontal plane.           & [0, 360] \\
    System Capacity    & kW             & Total DC power output of the system.                    & [1, 20] \\
    DC-to-AC Ratio     & -              & Ratio of DC to AC power capacity for the inverter.         & [1, 1.5] \\
    \bottomrule
    \end{tabular}
    \caption{Input variables for the Solar Energy Production (4D) experiment.}
    \label{tab:solar_energy_inputs}
\end{table*}

\section{Benchmarks}
\subsection{Levy Function (10D)}
The Levy function~\cite{c:SyntheticFunctions} is a popular synthetic function used to test global optimization methods, particularly in high-dimensional black-box functions. We restrict its search space to \([-10, 10]^{10}\) with a global maximum realized at $f(x) = 806.789$ when $x = (1, 1, \ldots, 1)$.

The 2D landscape of the Levy function is shown in Figure~\ref{fig:levy}. Uncovering its global maximum is challenging because of the rugged and oscillatory nature of its landscape with a large number of local minima, which increases with dimensionality. This function was selected to evaluate BORA’s capability to navigate high-dimensional, non-convex landscapes with multiple basins of attraction.

\subsection{Ackley Function (15D)}
\begin{figure}[!t]
    \centering
    \includegraphics[width=\columnwidth]{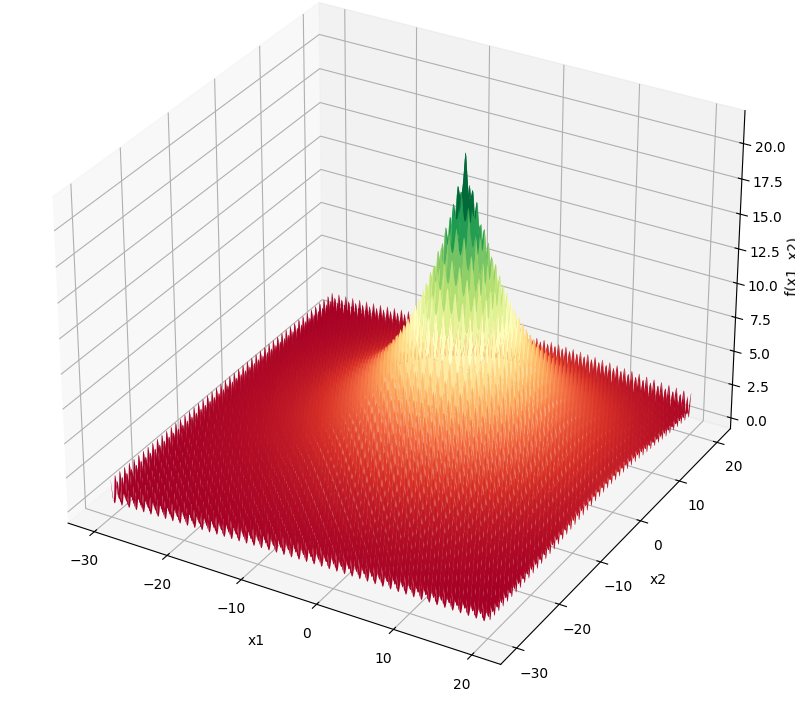}
    \caption{Landscape of the 2D Ackley function}
    \label{fig:ackley}
\end{figure}

The Ackley function~\cite{c:SyntheticFunctions} is another well-known synthetic benchmark, known for its flat outer region, steep gradients near the global minimum, many local minima, and symmetry. We changed the bound to $x \in [-30, 20]^d$, breaking its original symmetry. The global maximum is achieved at \(f(x) = 21.945\), when $x = (0, 0, \ldots, 0)$. The 2D landscape of the Ackley function is shown in Figure~\ref{fig:ackley}, highlighting the optimization challenge that results from the interplay between its wide, flat regions and highly oscillatory behavior with many local optima. Consequently, this can mislead optimization strategies that focus solely on exploration or exploitation. This function was selected to evaluate BORA’s capability to locate the global maximum efficiently while maintaining robustness in its search strategy.

\subsection{Solar Energy Production (4D)}
The Solar Energy Production experiment simulates the energy output of a solar panel system on January 1\textsuperscript{st} 2024, located at 35°N latitude, -120°W longitude (California) over a single day. The actual Californian weather conditions are programmatically retrieved, and the simulation is performed using the Python PVLib library for photovoltaic modeling~\cite{c:pvlib}. The experiment is described as follows:
\begin{itemize}
    \item \textbf{Objective}: Maximize the energy produced by the solar panel over a single day (24 hours).
    \item \textbf{Optimization Variables}: The input variables are the solar panel's tilt angle, azimuth angle, system capacity, and DC-to-AC ratio. They are described in detail in Table~\ref{tab:solar_energy_inputs}.
    \item \textbf{Target}: The daily energy output in kWh.
\end{itemize}

This experiment presents several challenges:
\begin{itemize}
    \item Multi-Dimensional Non-Convexity: The tilt and azimuth angles non-linearly impact the energy output, creating multiple local optima.
    \item System Trade-Offs: Balancing the system capacity and DC-to-AC ratio is essential to ensure efficiency and high energy output. 
    \item Day-Long Variability: As the simulation also accounts for daily variations in solar irradiance, the problem becomes time-dependent.
\end{itemize}
This problem was chosen to evaluate BORA's ability to optimize real-world engineering systems with dynamic dependencies, where practical knowledge is essential.

\subsection{Pétanque (7D)}
\begin{table*}[!t]
    \centering
    \begin{tabular}{l|l|lr}
        \toprule
        Variable & Unit & Description & Bound \\
        \midrule
        Pitch Angle                   & °              & Vertical launch angle relative to the ground where 0° is parallel to the ground.       & [-30, 90] \\
        Yaw Angle                     & °              & Horizontal aiming angle relative to the target where 0° is aiming straight.            & [-180, 180] \\
        Initial Velocity              & m/s            & Magnitude of the initial velocity.                                                     & [0, 50] \\
        Spin Rate                     & rpm            & Rotational speed of the projectile.                                                    & [0, 3000] \\
        Spin Axis Orientation         & °              & Direction of the spin axis in the xy-plane where 0° aligns with the +x-axis.           & [-180, 180] \\
        Initial Height (h)            & m              & Vertical height above ground level from which the ball is launched.              & [0, 2] \\
        Mass                          & kg             & Mass of the ball.                                                                      & [0.01, 10] \\
        \bottomrule
    \end{tabular}
    \caption{Input variables for the Pétanque (7D) experiment.}
    \label{tab:petanque_inputs}
\end{table*}

\begin{table*}[!t]
    \centering

    \begin{tabular}{lllr}
        \toprule
        Variable & Unit & Description & Bounds \\
        \midrule
        Irradiance       & kJ/m²/day     & Daily solar radiation received by the crop.       & [0, 8000] \\
        Minimum Temperature & °C         & Daily minimum temperature.                    & [-30, 30] \\
        Maximum Temperature & °C         & Daily maximum temperature.                    & [-20, 50] \\
        Vapor Pressure   & kPa           & Daily mean atmospheric vapor pressure.            & [0.06, 10] \\
        Wind Speed       & m/s           & Daily mean wind speed at 2 meters above the surface.    & [0, 5] \\
        Rainfall         & mm            & Total daily precipitation.                        & [0, 20] \\
        Water Availability & cm          & Initial water available in the total soil profile.& [0, 100] \\
        Soil Moisture Content & -         & Initial maximum moisture content in the rooting zone. & [0, 1] \\
        \bottomrule
    \end{tabular}
    \caption{Input variables for the Sugar Beet Production (8D) experiment.}
    \label{tab:sugar_beet_inputs}
\end{table*}

The Pétanque experiment simulates a simplified version of the pétanque game. In pétanque, players score points by throwing or rolling balls as close as possible to a stationary, smaller ball target. Players can choose balls of different sizes or weights depending on their strategy. In our Pétanque experiment, there is just a single player whose goal is to maximize their score by throwing a single ball closer to the fixed target located at $(x_t,y_t,z_t) = (50 \text{ meters}, 0, 0)$. Moreover, the ball is always launched from the initial position $(x_s, y_s, z_s)=(0,0,h)$ where $h$ is the vertical height from which the ball throw is released. The trajectory of the ball is tracked until it lands on the ground at $(x_e, y_e, z_e)$. We simulate the trajectory using Python with a $0.01$ time step following projectile motion equations, including factors like gravity, drag, and the Magnus effect for spinning projectiles. We record the distance between where the ball landed and the target as `distance\textunderscore{error}' and define the score ranging from 0 to 100 accordingly:
\begin{align}
    \text{distance\_error} &= \sqrt{(x_e-x_s)^2+(y_e-y_s)^2+(z_e-z_s)^2} \nonumber \\
    \text{score} &= 100 \cdot \exp(-\text{distance\_error} \cdot 0.2)
\end{align}
In our setup, better accuracy corresponds to a smaller distance error and, thus, a higher score. The experiment is described as follows:
\begin{itemize}
    \item \textbf{Objective}: Maximize the accuracy of the throw to maximize the score.
    \item \textbf{Optimization Variables}: The launch variables include initial velocity, angles, spin, and ball mass, further described in detail in Table~\ref{tab:petanque_inputs}.
    \item \textbf{Target}: The score.
\end{itemize}

The Pétanque experiment presents several challenges:
\begin{itemize}
    \item Multi-dimensional Input Space: The throw involves seven input variables that create a high-dimensional search space.
    \item Non-Linear Dynamics: The trajectory of the path depends non-linearly on each input.
    \item Throw Accuracy: Finely tuned launch inputs are necessary to achieve high accuracy.
\end{itemize}
This problem was chosen to evaluate BORA's ability to optimize physical systems under complex dynamics.

\begin{table*}[!t]
    \centering
    \begin{tabular}{lllrr}
        \toprule
        Variable                        & Unit          & Description                               & Bound     & Discretization Step\\
        \midrule
        Acid Red 871                    & mL            & A dye.                                    & [0, 5]    & 0.25\\
        L-Cysteine                      & mL            & Hole scavenger.                           & [0, 5]    & 0.25\\
        Methylene Blue                  & mL            & A dye.                                    & [0, 5]    & 0.25\\
        Sodium Chloride                 & mL            & Adjusts ionic strength of the solution.   & [0, 5]    & 0.25\\
        Sodium Hydroxide                & mL            & Controls pH of the reaction.              & [0, 5]    & 0.25\\
        P10-MIX1                        & g             & A conjugated polymer photocatalyst.       & [1, 5]    & 0.20\\
        Polyvinylpyrrolidone (PVP)      & mL            & A surfactant.                             & [0, 5]    & 0.25\\
        Rhodamine B                     & mL            & A dye.                                    & [0, 5]    & 0.25\\
        Sodium Dodecyl Sulfate (SDS)    & mL            & A surfactant.                             & [0, 5]    & 0.25\\
        Sodium Disilicate               & mL            & Stabilizes the reaction medium.           & [0, 5]    & 0.25\\
        \bottomrule
    \end{tabular}
    \caption{Input variables for the Hydrogen Production (10D) experiment.}
    \label{tab:photocatalytic_inputs}
\end{table*}

\subsection{Sugar Beet Production (8D)}
Sugar beet is a plant with a high sucrose concentration in its root, making it commercially useful for raw sugar production. The Sugar Beet Production experiment examines its production in greenhouses, which allows a high level of control over the weather and soil conditions. We use the World Food Studies Simulation model (WOFOST)~\cite{c:WOFOST}, available within the Python Crop Simulation Environment (PCSE)~\cite{c:PCSE}, to simulate the growth of sugar beets in a controlled greenhouse environment. WOFOST is an explanatory model capable of tracking crop growth over daily time steps, integrating a detailed understanding of physiological and environmental processes. The experiment is described as follows:
\begin{itemize}
    \item \textbf{Objective}: Maximize the total aboveground biomass (TAGP) of sugar beet crops in a greenhouse environment over a month (31-day) period. The initial sugar beet biomass is 0.41 kg/ha.
    \item \textbf{Optimization Variables}: The input variables are the greenhouse weather and soil conditions, which are further detailed in Table~\ref{tab:sugar_beet_inputs}. These variables remain constant during the month simulation period.
    \item \textbf{Target}: TAGP at the end of the simulation in kg/ha.
\end{itemize}

The Sugar Beet Production experiment presents several challenges:
\begin{itemize}
    \item Inter-dependencies: Variables like moisture, irradiance, and temperatures are interdependent and significantly influence crop growth.
    \item Sensitivity: The growth simulation is notably sensitive to variables such as soil moisture content and vapor pressure, complicating the optimization process.
\end{itemize}

This problem was chosen to evaluate BORA's ability to optimize agricultural yields in complex, sensitive, and non-linear systems.

\subsection{Hydrogen Production (10D)}
\label{s:hydrogen_production}
The Hydrogen Production experiment was orignally realized by~\citeauthor{c:MobileRoboticChemist} to find the optimal composition of a ten-component mixture maximizing the Hydrogen Evolution Rate (HER) via photocatalysis. Due to physical constraints in the experiment, ~\citeauthor{c:MobileRoboticChemist} constrained the search space to keep the total mixture volume less than 5 mL (the maximum volume in the reaction vial). Other constraints come from the liquid dispensing module (minimum accurate dispense = 0.25 mL) and the solid dispensing module (minimum dispense = 0.2 mg), turning the idealized continuous space into a discretized one. The resulting combinatorial search space was extremely large, with 98,423,325 possible combinations,~\citeauthor{c:MobileRoboticChemist}, so the authors used an autonomous mobile robotic chemist along with BO to search for the optimal composition of the chemical mixture.

To replicate this experiment, we used the same cost-effective and time-saving method as~\citeauthor{c:HypBO}, who are part of the same research group as ~\citeauthor{c:MobileRoboticChemist}.~\citeauthor{c:HypBO} built an oracle method by training a Gaussian process regression on a dataset of 1119 experimental observations of the original study to serve as `ground truth', or oracle model. Although this interpolated model is an approximation, ~\citeauthor{c:HypBO} demonstrated in their HypBO work that it captured the underlying chemical processes of the original experiment, making it reliable enough to draw safe conclusions.

The Hydrogen Production experiment details are as follows:
\begin{itemize}
    \item \textbf{Objective}: Maximize the HER ($\mu$mol/h), by optimizing the quantities of each of the ten chemicals in the mixture.
    \item \textbf{Optimization Variables}: The input variables represent the amount of each chemical in the mixture. They are described in detail in Table~\ref{tab:photocatalytic_inputs}.
    \item \textbf{Constraints}: To ensure compatibility with the experimental setup,
        \begin{enumerate}
            \item Total Concentration: The sum of all liquid chemicals (excluding P10-MIX1, which is solid) must not exceed 5.
            \item Discrete Variables: All input variables are discretized (see~\ref{tab:photocatalytic_inputs}).
        \end{enumerate}
    \item \textbf{Target}: The HER predicted by the oracle model.
\end{itemize}

This experiment presents several challenges:
\begin{itemize}
    \item High-Dimensional Search Space: The optimization involves 10 discrete parameters, resulting in a large combinatorial experimental space.
    \item Complex Interactions: HER depends on complex non-linear chemistry interplay among the chemical inputs.
    \item Physical Constraints: The constraint in the mixture volume imposes an additional challenge for the optimization process.
\end{itemize}

This problem was chosen to evaluate BORA's ability to optimize complex, high-dimensional, constrained chemistry problems.

\section{Baselines}
\begin{figure*}[!t]
    \centering
    \includegraphics[width=\textwidth]{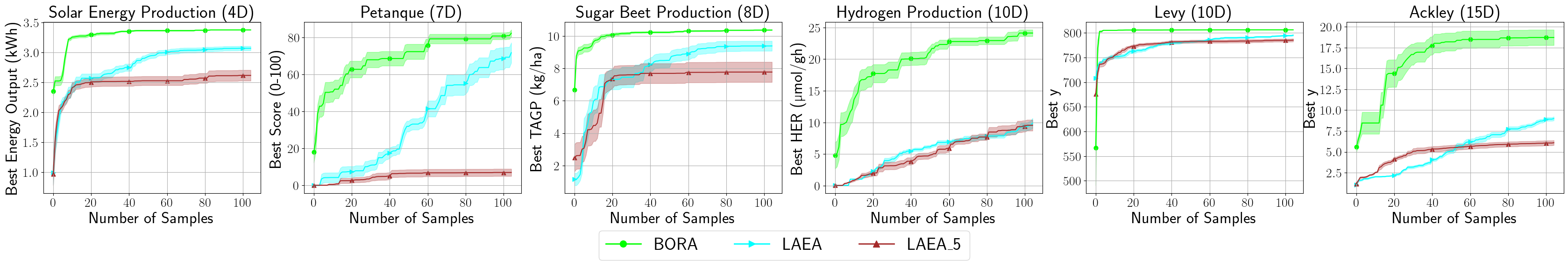}
    \caption{LAEA vs. LAEA\_5. LAEA denotes LAEA with an initial population size of 15. Solid lines show mean values while shaded areas indicate $\pm 0.25$ standard error.}
    \label{fig:laea_5_vs_15}
\end{figure*}

All of the baselines that we tested BORA against have internal random sampling mechanisms to generate random points. In the BO-based baselines, these random points are partially used as candidate points for the acquisition function. In the LAEA method, they contribute to the generation and evolution of the population during the optimization. To enable a fair comparison between BORA and baseline methods for the Hydrogen Production experiment, we adapted the random sampling mechanism of each baseline to effectively handle the discrete and constrained nature of the input space. This ensures that the baselines remained competitive and adhered to the unique constraints of the Hydrogen Production problem. Note that HypBO did not need modification as it inherently supports such spaces.

This adaptation was achieved by first employing a simplex-based approach to scale randomly generated points to satisfy the sum constraint. Then, we employed a discretization step to ensure that the generated points align with the problem's discrete requirement. Additionally, a caching system was implemented to exclude duplicate points, enhancing diversity and avoiding redundancies. Aside from the modifications made to the random sampler to account for the discrete and constrained nature of the Hydrogen Production experiment, the baseline implementations were left unchanged and are consistent with their original formulations. All default hyperparameters were used as specified in their respective papers.

\subsection{LAEA Population Size}
For the LAEA baseline, we experimented with two initial population sizes: 5 and 15. An initial population size of 5 aligns with the 5 initial samples we use for all BORA and all the other baselines, ensuring a consistent starting point across methods. However, as LAEA is an LLM/Evolutionary Algorithm hybrid, it differs fundamentally from the other methods explored here, which are all BO-based except for Random Search. Consequently, the same rationale for selecting initial samples in BO does not necessarily extend to LAEA. To account for this, we also tested an initial population size of 15, which preserves a consistent population-to-optimization budget ratio as in the original paper~\cite{c:LAEA}. Our experiments, as illustrated in Figure~\ref{fig:laea_5_vs_15}, showed that LAEA, with an initial population size of 15, outperformed the smaller population size of 5. Consequently, we report results for LAEA 15 in the manuscript to present its best performance.

\subsection{ColaBO Input Priors by an LLM}
As stated in the manuscript, to avoid the impracticality of relying on humans to provide inputs for multiple trials across all experiments for ColaBO, we used the LLM GPT-4o-mini to generate the `human' inputs. It is likely, of course, that human domain experts will in some cases provide superior inputs than produced by an LLM, and we note that limitation here. ColaBO uses GP priors to express human input as:
\[
    \pi(x) \sim \mathcal{N}(c,\sigma),
\]
where $c$ is the center of the prior and speculated to be the optimum, and $\sigma$ is the standard deviation.

To automate the production of the input prior, we used the same task description prompts as used for BORA and tasked an LLM to guess the location of the optimum with a confidence score ranging between 0 and 1. The given location becomes the center of the GP. The prior standard deviation is calculated to be
\[
    \sigma = 1 - \text{score} + \epsilon,
\]
where $\epsilon=10^{-6}$ is a small number to prevent a null standard deviation. In this setup, the more confident the LLM is in its guess, the narrower the input prior becomes.

\subsection{HypBO Hypotheses by an LLM}
Unlike in BORA, HypBO~\cite{c:HypBO} defines a hypothesis as a static region of interest; that is, a `box soft constraint'. We automated the generation of these hypotheses in testing HypBO on the benchmarks by using an LLM. Similar to the ColaBO input prior generation, we used LLM GPT-4o-mini and the same task description prompts as used for BORA. Again, the quality of the LLM input may not match that of a human domain expert, but this was done because of the impracticality of drawing on human experts over multiple trial optimization runs. The LLM is tasked to predict the location of the optimum and to give a [0, 1] confidence score. Because HypBO supports the use of multiple static hypotheses, we do not restrict the LLM in the number of hypotheses that it can recommend. Then, for each hypothesis, we use the corresponding score to `draw' a hyper-rectangle around its suggested optimum location. The length of the hyper-rectangle $L$ corresponds to the experiment space scaled using the confidence score as follows:
\[
    L = (\text{ub} - \text{lb}) * (1 - \text{score}),
\]
where ub and lb are vectors corresponding to the upper and lower bounds of the experiment input space. In this setup, the more confident the LLM is in its guess, the narrower the input prior becomes.

\subsubsection{Case of the Hydrogen Production Benchmark}
For the Hydrogen Production experiment, because it was originally optimized in the HypBO paper~\cite{c:HypBO} with some relevant hypotheses, instead of using an LLM to generate hypotheses for it, we employed the most realistic hypothesis used in that study and the corresponding optimization data. That hypothesis is `What They Knew', which encapsulates any human knowledge available prior to the execution of the experiment, thus precluding any post-experiment hindsight. As such, the LLM is used here to simulate the human input for HypBO for all tasks studied here, apart from Hydrogen Production, where the input is domain-expert human knowledge from~\cite{c:HypBO}.
 
\section{Extended Methodology}
As BORA is a hybrid method involving LLMs, its effectiveness depends, in part, on the quality and the structure of the LLM outputs. Additionally, ensuring that the LLM outputs respect a desired format guarantees that BORA is fully automated. This section details the LLM prompt engineering, including the reflection strategies used to build robust prompts for the LLM agent to:
\begin{itemize}
    \item Warm-start the optimization;
    \item Comment on the optimization progress and generate hypotheses when progress stalls via the actions $a_{2}$ and $a_{3}$;
    \item Summarize the optimization findings and recommend further experiments.
\end{itemize}

We also detail the fallback mechanisms in the event that the LLM fails to produce the desired output.

\subsection{LLM Role-Based Prompting}
Our prompt engineering starts with role-based prompting, which allows us to seed the context and behavior of the LLM agent at the start of the session by explicitly assigning it a role to ensure that it operates consistently within our framework. Our LLM agent is assigned the role of a research assistant specialized in BO and also that of a live commentator, commenting on the optimization progress of an experiment and generating hypotheses. This role prompting is enhanced with the \emph{Experiment Card} provided by the user to ensure that the LLM agent understands the problem at hand and how, specifically, it will participate in the optimization process. 

\subsubsection{Experiment Card}
\begin{figure}[h]
    \centering
    \includegraphics[width=\columnwidth]{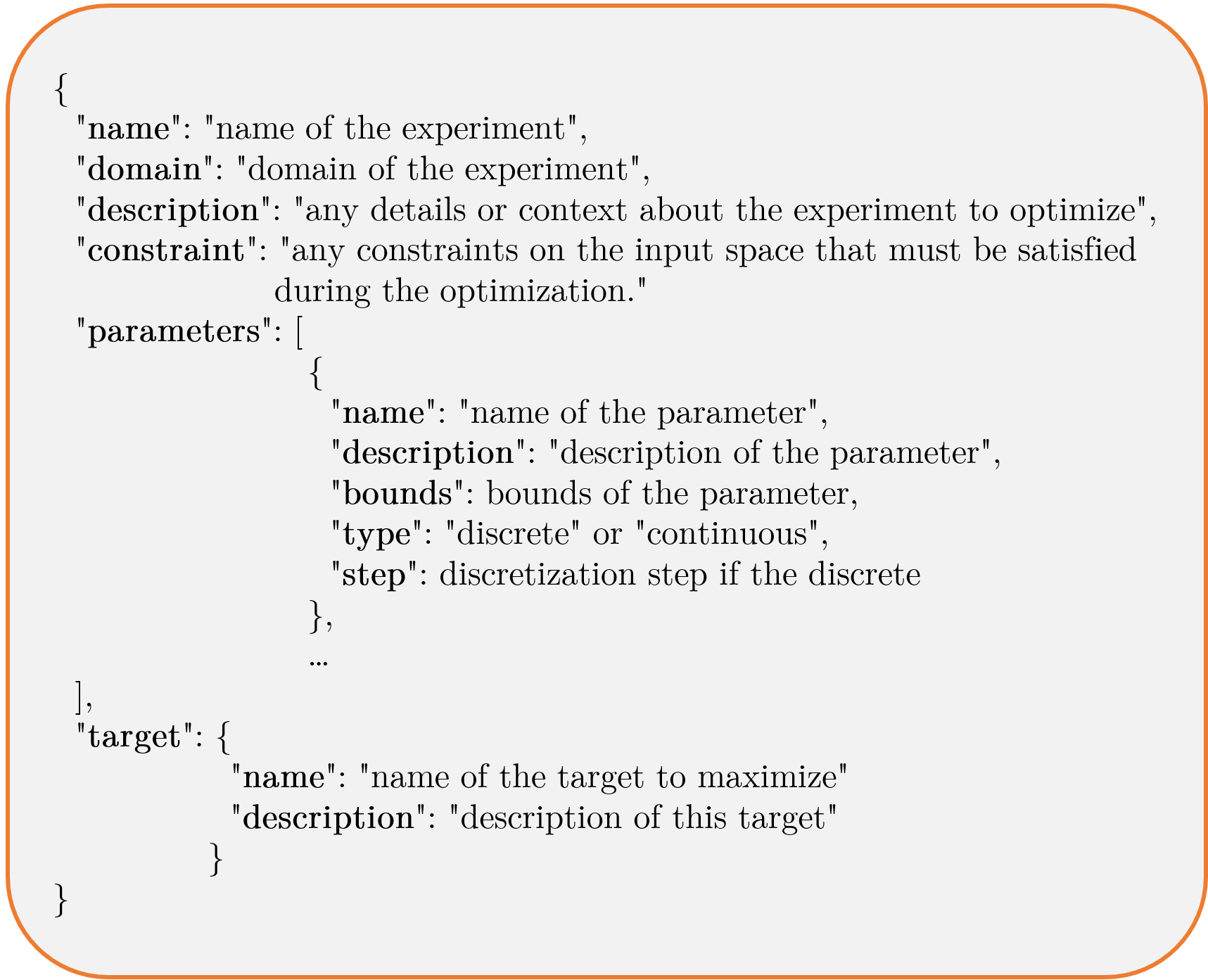}
    \caption{Experiment card template.}
    \label{fig:experiment_card_template}
\end{figure}

\begin{figure}[!t]
    \centering
    \includegraphics[width=\columnwidth]{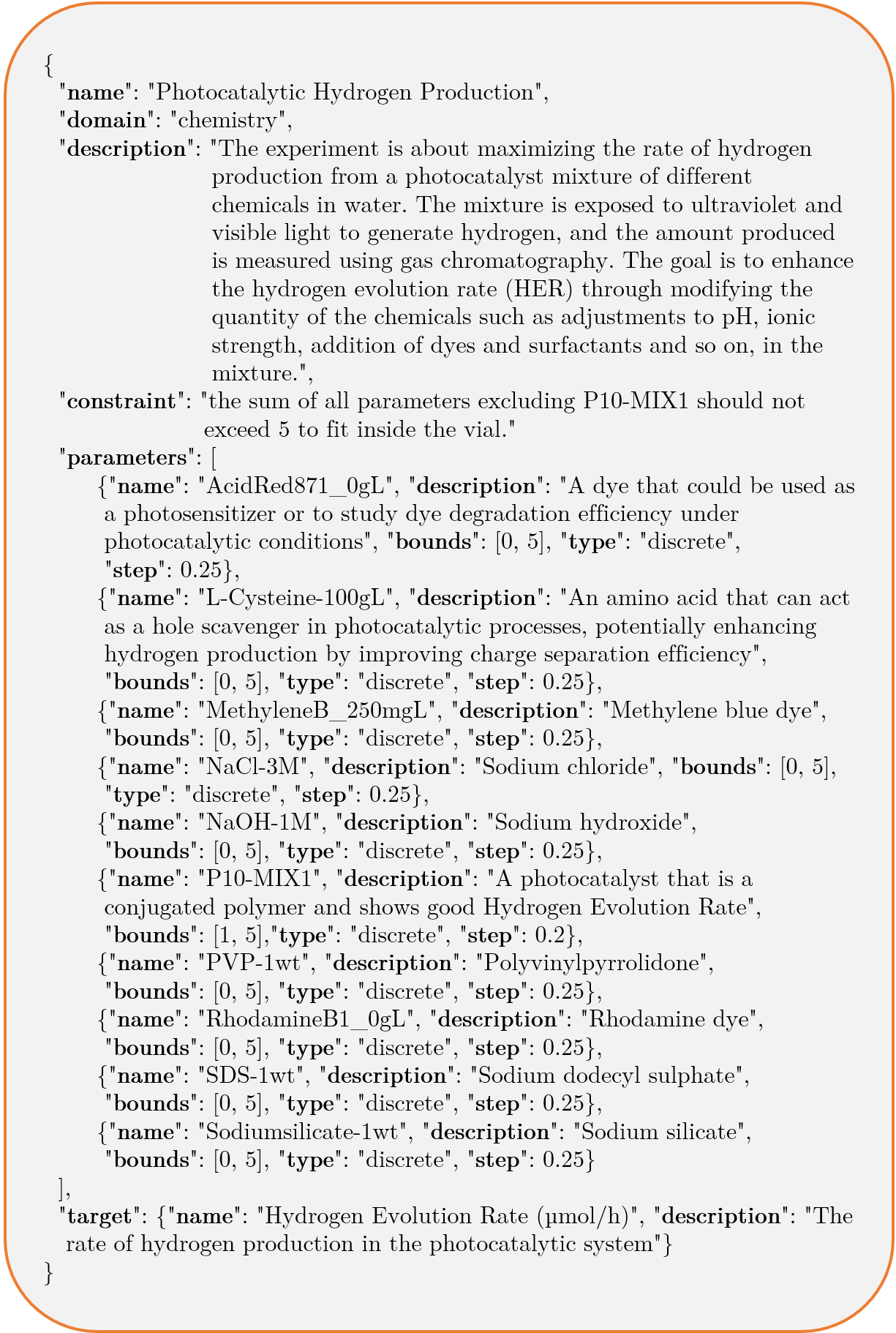}
    \caption{Experiment card of the Hydrogen Production experiment.}
    \label{fig:experiment_card_hydrogen_production}
\end{figure}

This card is constructed by the user following the standardized template shown in Figure~\ref{fig:experiment_card_template}. This template allows the construction of a comprehensive description of the experiment, helping the LLM contextualize the problem at hand. Figure~\ref{fig:experiment_card_hydrogen_production} shows what this looks like for the Hydrogen Production experiment.

\subsubsection{Enhanced Role Prompting}
\begin{figure}[!t]
    \centering
    \includegraphics[width=\columnwidth]{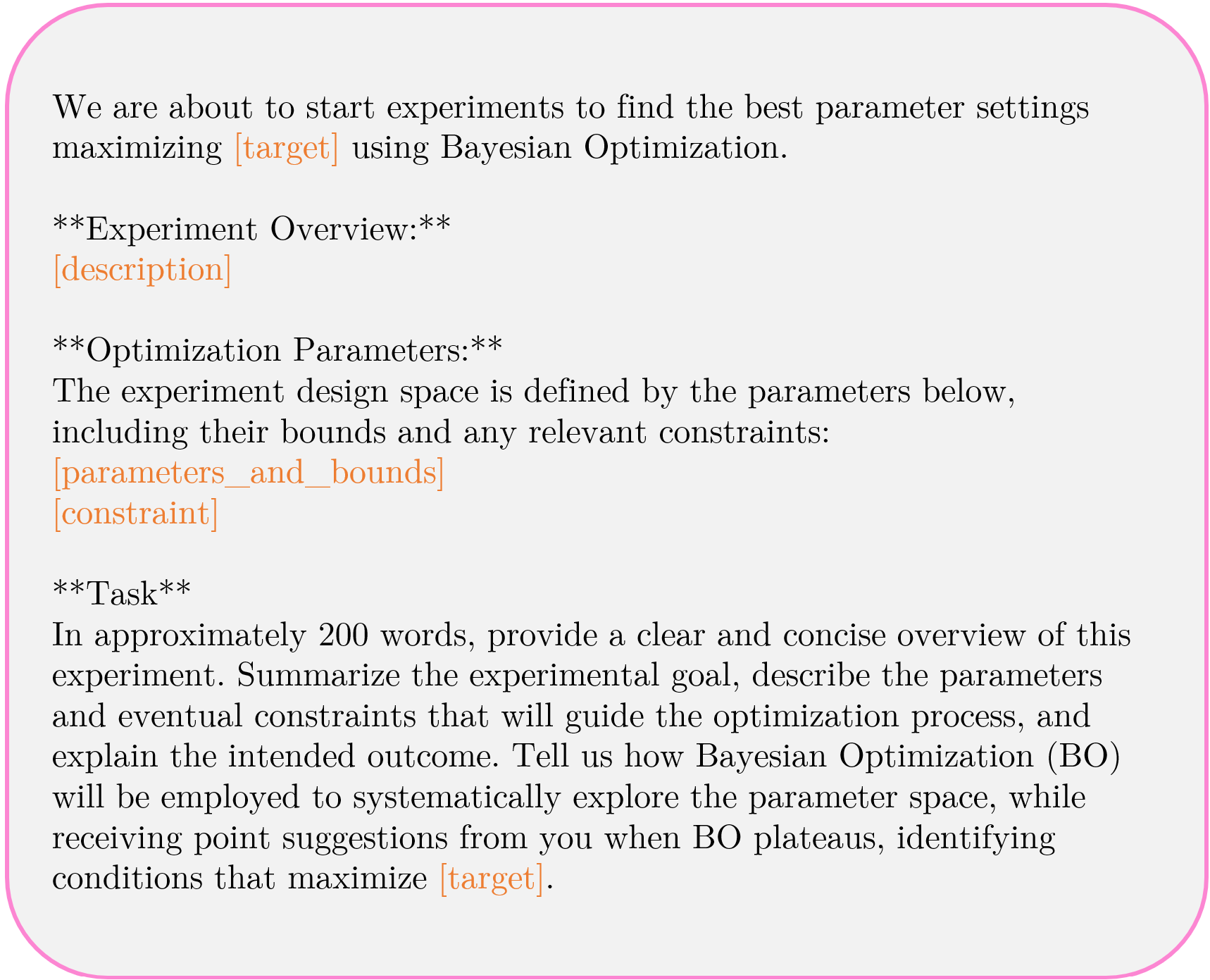}
    \caption{Prompt template tasking the LLM to give an overview of the experiment and to explain how it will participate in the optimization process.}
    \label{fig:experiment_overview_prompt_template}
\end{figure}

\begin{figure}[!t]
    \centering
    \includegraphics[width=0.9\columnwidth]{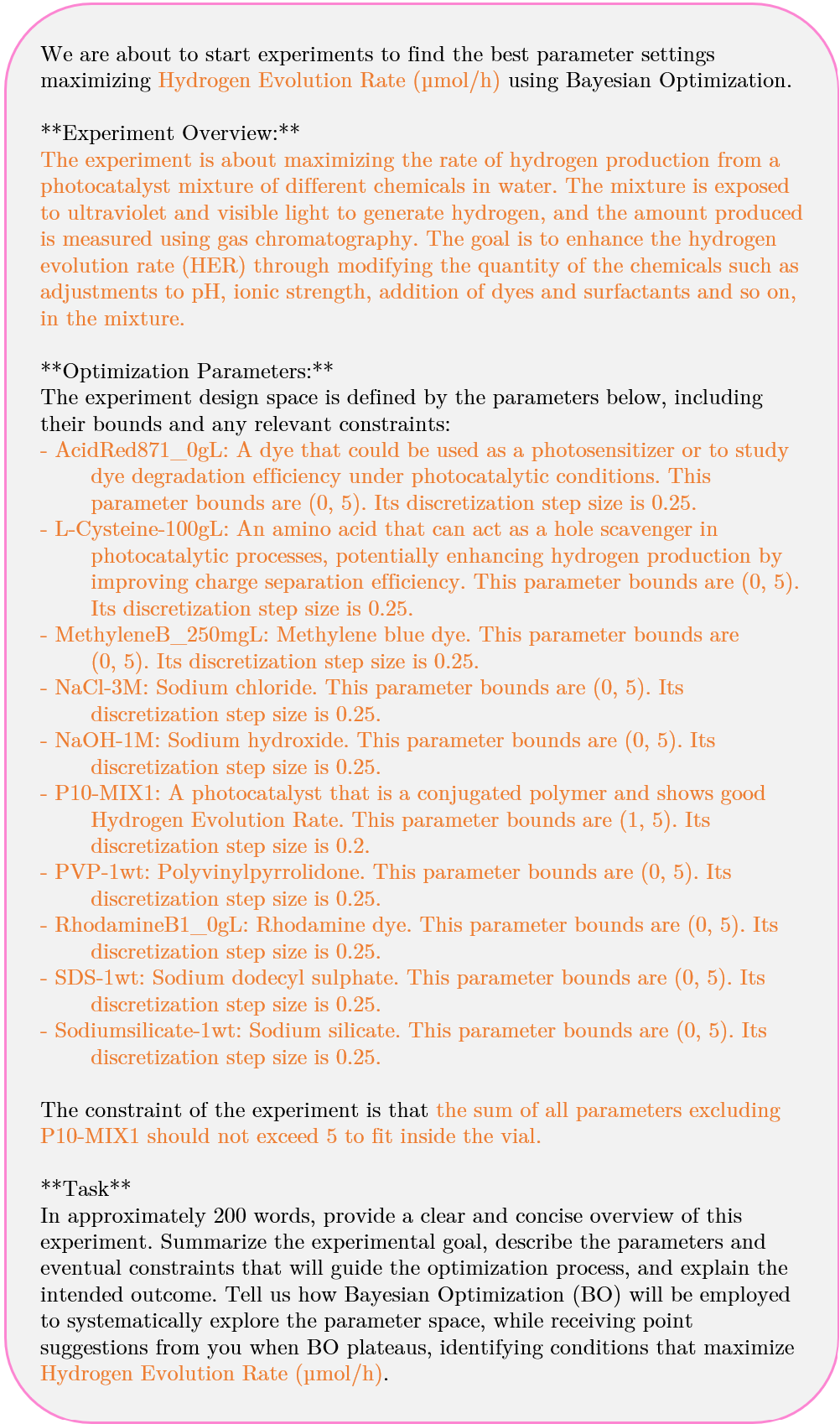}
    \caption{Prompt for the experiment overview given to the LLM agent about the Hydrogen Production experiment.}
    \label{fig:experiment_overview_prompt_hydrogen_production}
\end{figure}

This is achieved by leveraging the experiment card and prompting the LLM to give an initial overview of the experiment and to explain its role in the BORA optimization procedure. Figure~\ref{fig:experiment_overview_prompt_template} illustrates this prompt template, while Figure~\ref{fig:experiment_overview_prompt_hydrogen_production} exemplifies it for the Hydrogen Production experiment. Note that in the prompt figures, [] indicates a placeholder, while:
\begin{itemize}
    \item \textcolor{orange}{Orange} text corresponds to information retrieved from the experiment card.
    \item \textcolor{green}{Green} text corresponds to specific instructions.
    \item \textcolor{blue}{Blue} text corresponds to information about or the gathered data itself.
    \item \textcolor{purple}{Purple} text corresponds to any previous comments made by the LLM.
\end{itemize}

\begin{figure}[!t]
    \centering
    \includegraphics[width=0.93\columnwidth]{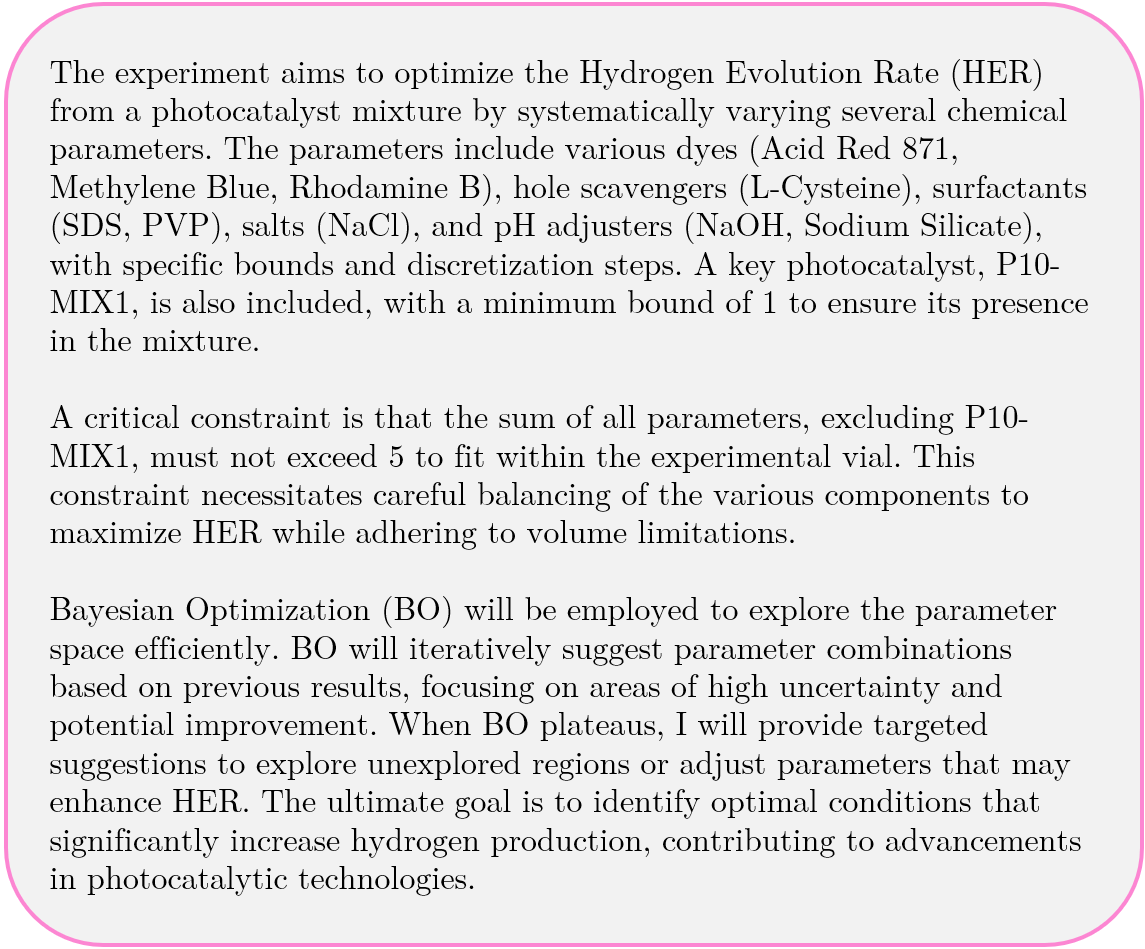}
    \caption{Example of the LLM's overview of the Hydrogen Production experiment.}
    \label{fig:experiment_overview_hydrogen_production}
\end{figure}

The output of this role-based prompting (Figure~\ref{fig:experiment_overview_hydrogen_production}) is also returned to the user and marks the beginning of BORA's narrative framework. Beyond ensuring that the LLM agent is properly initialized, it also fosters transparency and trust between BORA and the user by:
\begin{itemize}
    \item Confirming to the user that BORA understands the problem given by the user.
    \item Explaining to the user how BORA works by having an agent assisting BO in its optimization procedure.
\end{itemize}

\newpage
\subsection{Comment Object}
\label{s:comment}
Beyond the role-based prompting, the LLM agent is prompt-engineered to return a specific JSON object referred to as \emph{Comment} during its interventions in the optimization process. This structure is illustrated in Figure~\ref{fig:initialization_prompt_template} and contains:
\begin{itemize}
    \item Insights: Comments on optimization progress and key findings. Note that while in the Comment object structure shown in Figure~\ref{fig:initialization_prompt_template} this corresponds to \emph{comment}, for the sake of clarity and to avoid confusion in this manuscript, we refer to it as \emph{insights}. 
    \item Hypotheses: A list of hypotheses maximizing the target, each accompanied by:
        \begin{itemize}
            \item A meaningful name.
            \item A rationale defending the hypothesis.
            \item A confidence level indicating the LLM confidence in the hypothesis.
            \item A list of points to test the hypothesis.
        \end{itemize}
\end{itemize}

Note that, as we will see in the following sections, while the \textit{Comment }structure was designed to be general, we only ask the LLM for one point per hypothesis. This was based on the ablation studies in HypBO~\cite{c:HypBO}, which showed that a well-selected point in a hypothesis region can transfer enough information to BO. Additionally, this limits the LLM consumption of its optimization budget. Moreover, the confidence level is for the user to get a sense of how confident the LLM is in its recommendations. The value placeholders are automatically generated by randomly sampling points within the experiment design space based on the Experiment Card information.

\subsection{Optimization Initialization}
\label{s:warm_start}
\begin{figure}[!t]
    \centering
    \includegraphics[width=\columnwidth]{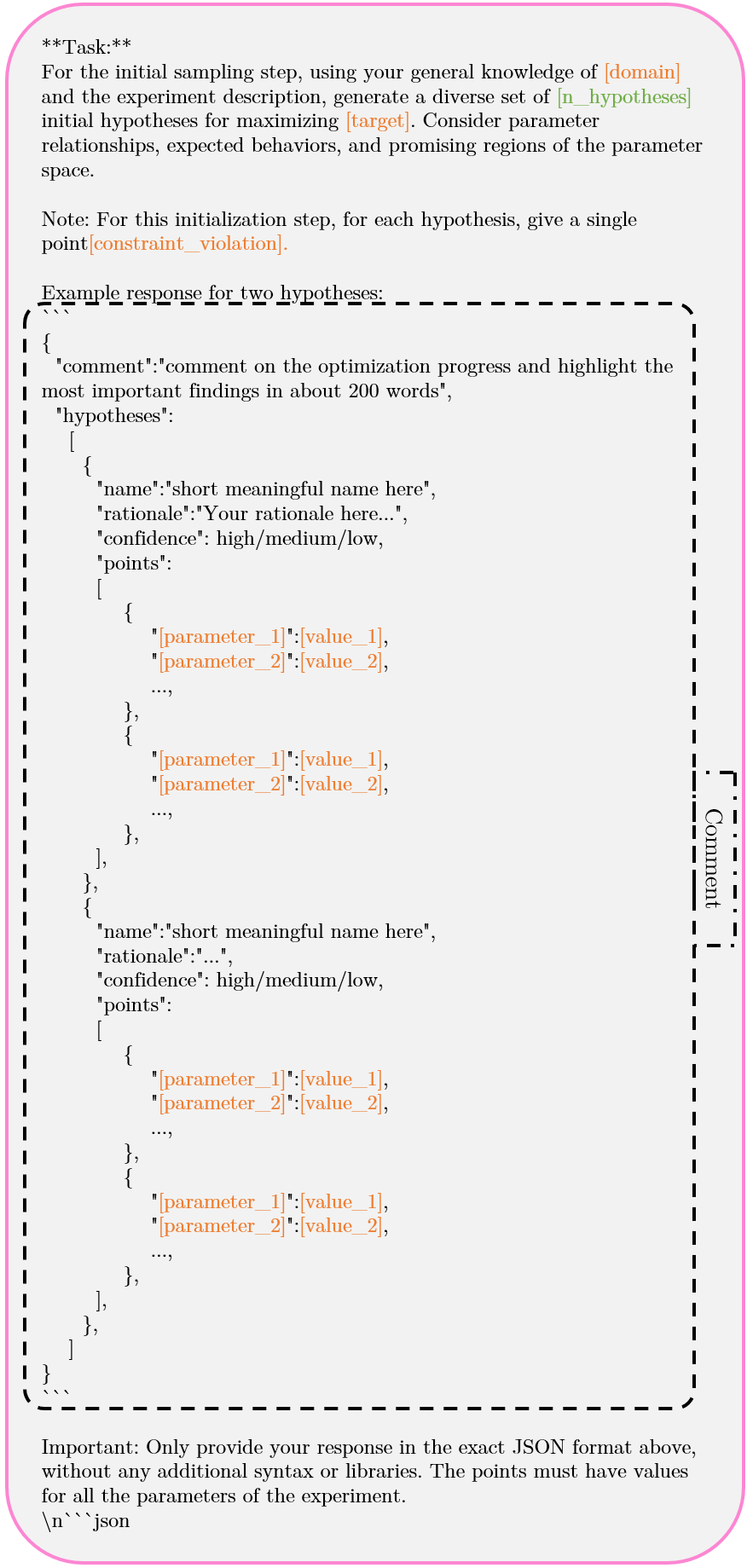}
    \caption{Prompt template to generate the initial samples.}
    \label{fig:initialization_prompt_template}
\end{figure}

\begin{figure}[!t]
    \centering
    \includegraphics[width=\columnwidth]{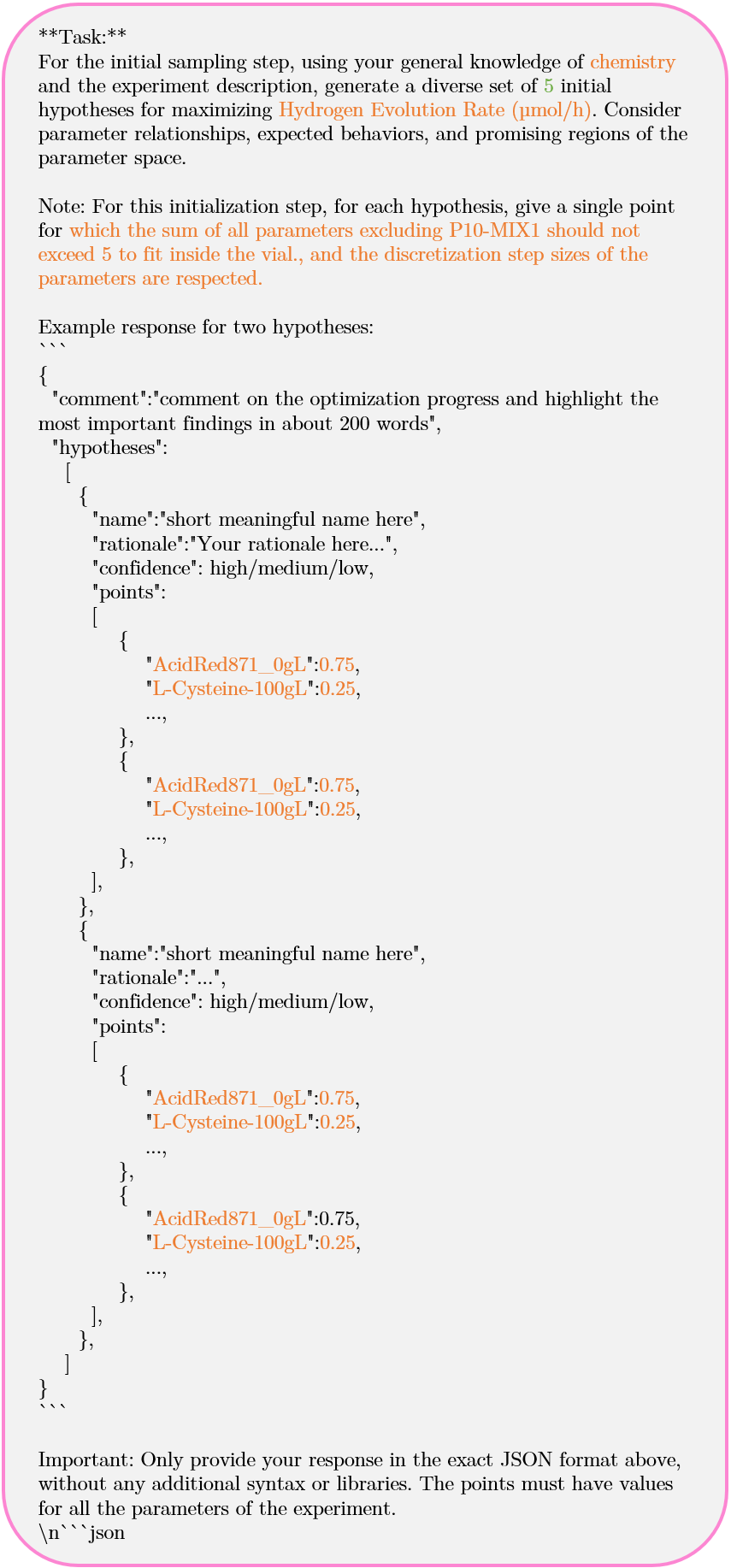}
    \caption{Example of the LLM prompt to initialize the Hydrogen Production experiment.}
    \label{fig:initialization_prompt_hydrogen_production}
\end{figure}

\begin{figure}[!t]
    \centering
    \includegraphics[width=\columnwidth]{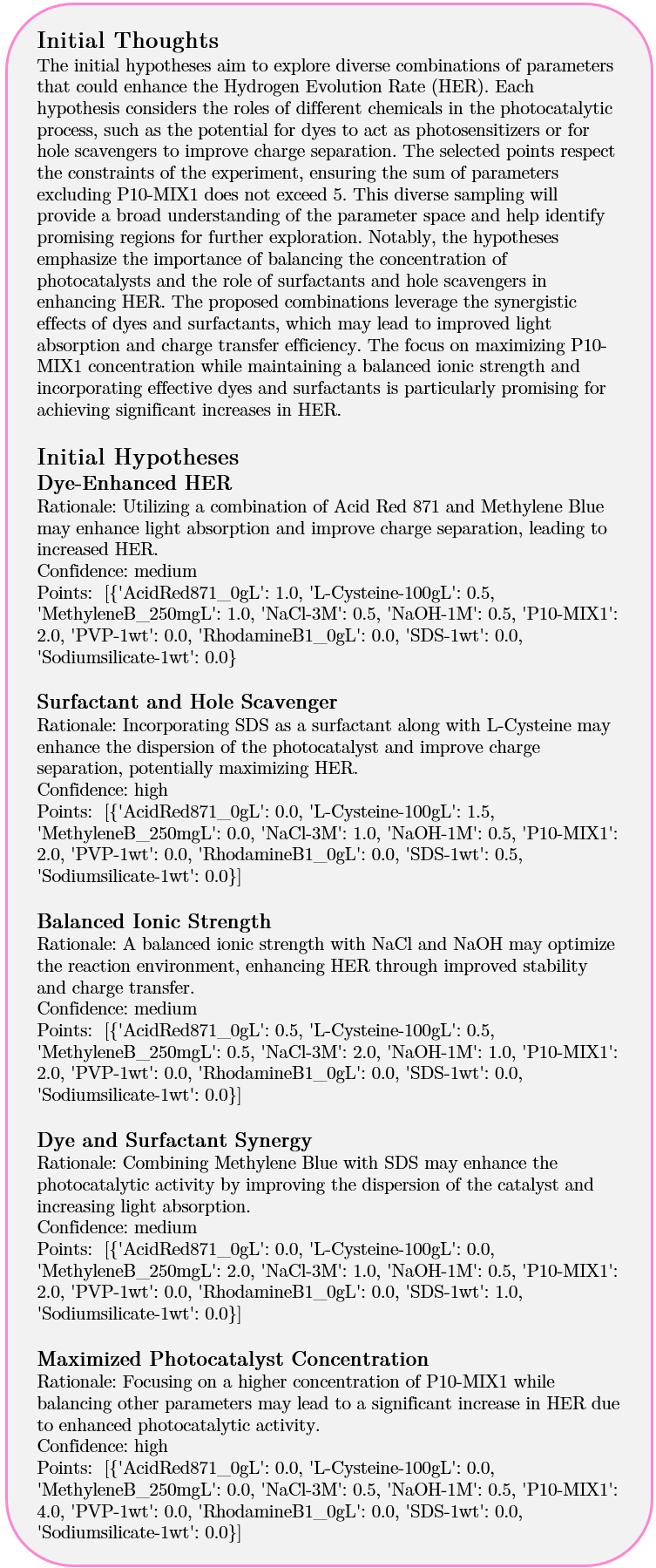}
    \caption{Example of the LLM initializing the Hydrogen Production experiment}
    \label{fig:initialization_hydrogen_production}
\end{figure}

The LLM warm-starts the optimization using any domain knowledge of the problem at hand. It is tasked to generate $n_{\text{hypotheses}} \gets n_{\text{init}}$ diverse hypotheses maximizing the target. Each of these hypotheses returns a single point that will be later evaluated to form the initial dataset. In this context, the LLM is few-shot prompted with the Comment structure described above and shown in the \emph{Initialization Prompt} Figure~\ref{fig:initialization_prompt_template}. The latter details our few-shot prompting technique leveraging direct instructions, formatting guidelines, and demonstrations via In-Context Learning (ICL) to make the LLM return a Comment object that BORA can consume. An example of the constructed initialization prompt for the Hydrogen Production experiment is detailed in Figure~\ref{fig:initialization_prompt_hydrogen_production}. Note that while the structured LLM outputs are consumed by BORA, these outputs are reformatted before being presented to the user. For readability, we are showing these user-friendly outputs instead of the raw LLM structured outputs. Figure~\ref{fig:initialization_hydrogen_production} illustrates what this initial comment looks like for the initialization of the Hydrogen Production experiment.

\subsection{LLM Interventions}
\label{s:llm_interventions}
\subsubsection{$a_{2}$ LLM Comments and Suggests $n_{\text{LLM}}$ Points}
\begin{figure}[h]
    \centering
    \includegraphics[width=\columnwidth]{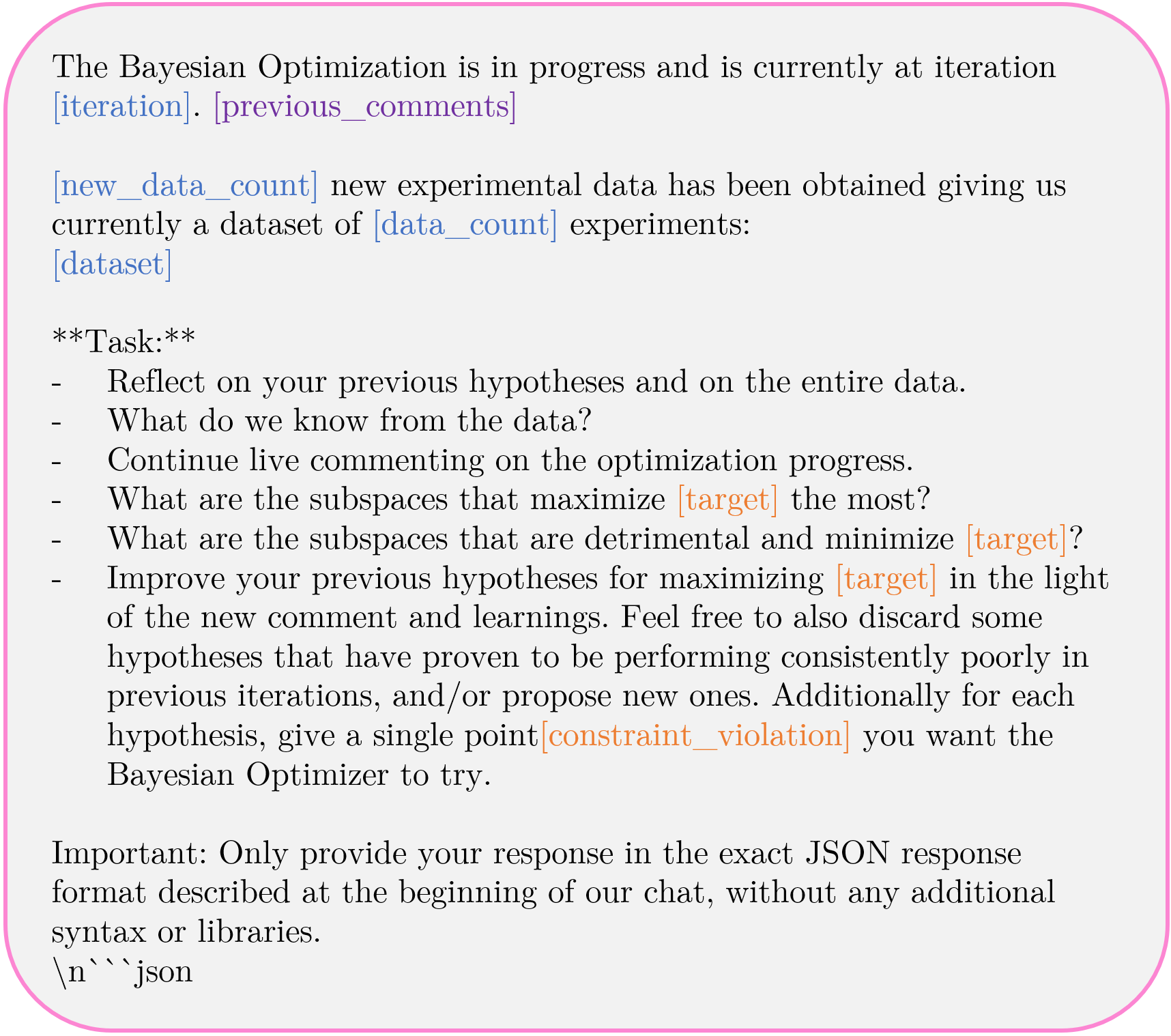}
    \caption{Prompt template tasking the LLM to to perform action $a_{2}$ (comment and suggest the next points to evaluate).}
    \label{fig:comment_and_suggest_prompt_template}
\end{figure}

The LLM is prompted following the template shown in Figure~\ref{fig:comment_and_suggest_prompt_template} with the gathered data up to the current iteration and any previous comments. Then, it is tasked to comment on the optimization progress in light of the new data and to update any previous hypotheses.

\begin{figure}[!t]
    \centering
    \includegraphics[width=\columnwidth]{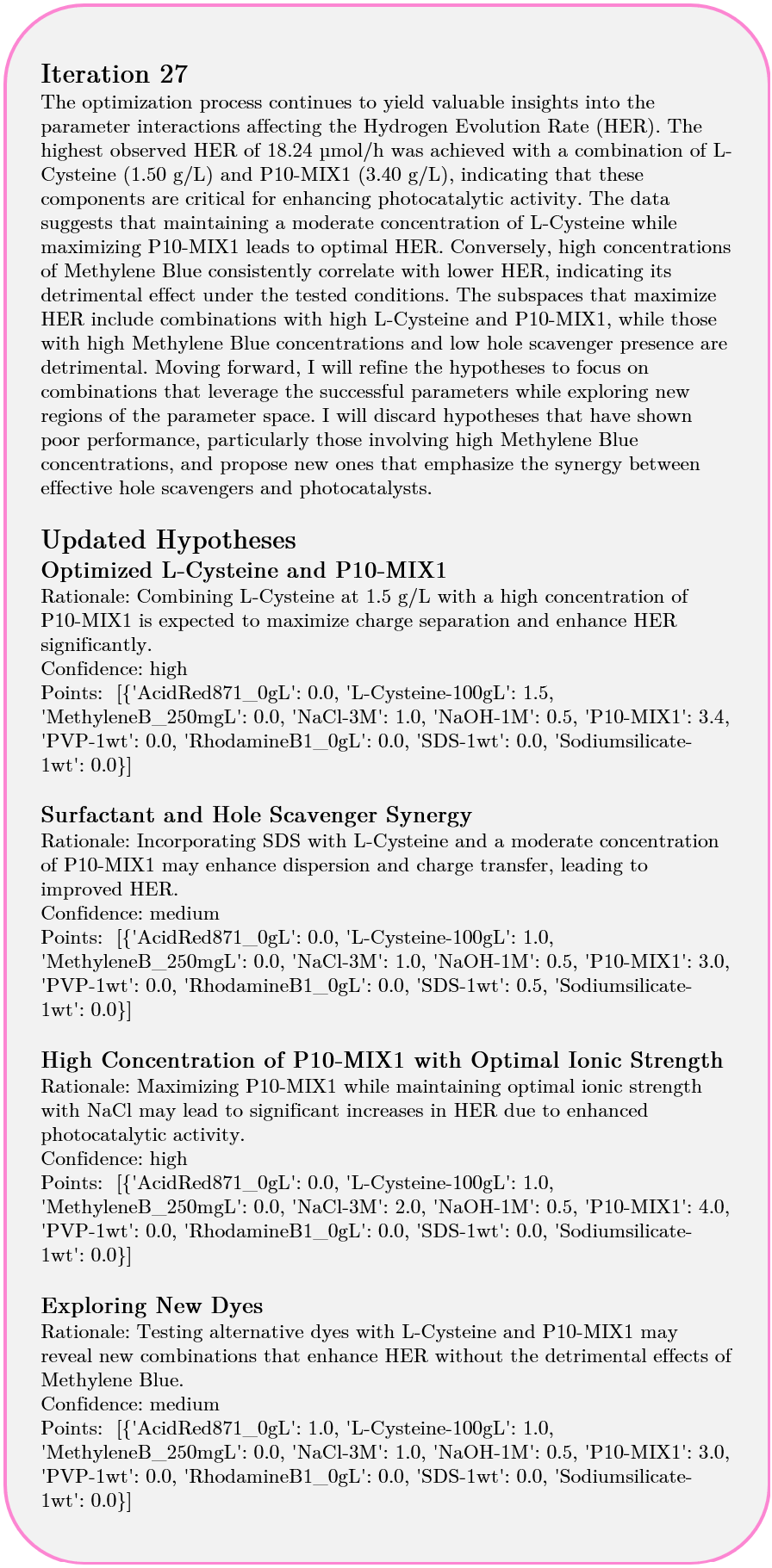}
    \caption{Example of the LLM performing action $a_{2}$ (comment and suggest the next points to evaluate) for the Hydrogen Production experiment.}
    \label{fig:comment_and_suggest_hydrogen_production}
\end{figure}

\begin{figure}[!t]
    \centering
    \includegraphics[width=\columnwidth]{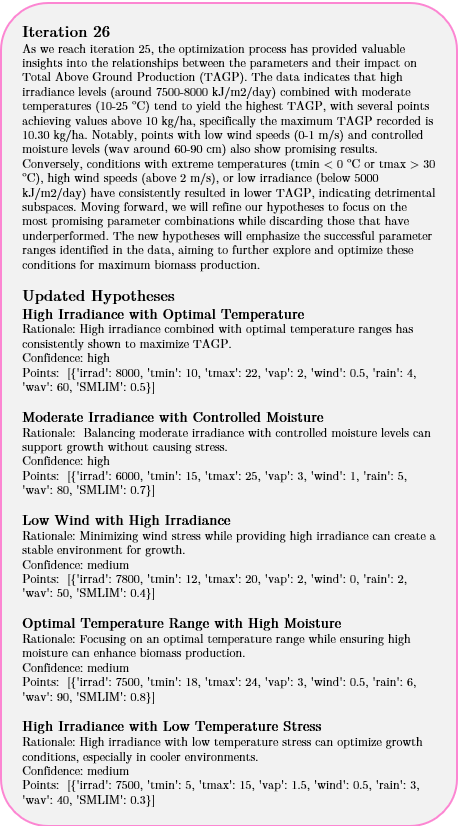}
    \caption{Example of the LLM performing action $a_{2}$ (comment and suggest the next points to evaluate) for the Sugar Beet Production experiment.}
    \label{fig:comment_and_suggest_sugar_beet}
\end{figure}

\begin{figure*}[!t]
    \centering
    \includegraphics[width=0.96\textwidth]{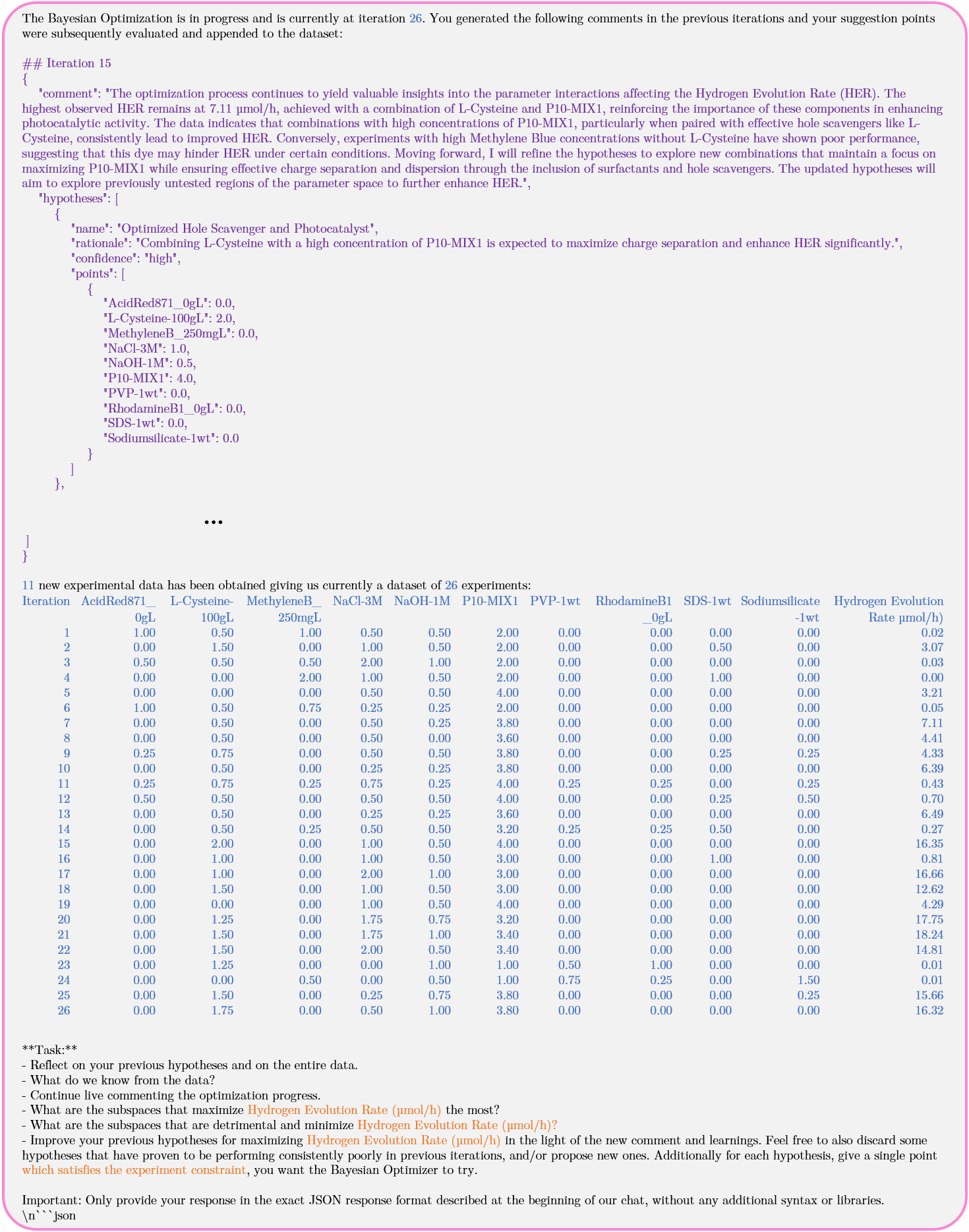}
    \caption{Example of the prompt received by the LLM to perform action $a_{2}$ (comment and suggest the next points to evaluate) for the Hydrogen Production experiment.}
    \label{fig:comment_and_suggest_prompt_hydrogen_production}
\end{figure*}

We show in Figure~\ref{fig:comment_and_suggest_prompt_hydrogen_production} an example of the constructed prompt received by the LLM for the Hydrogen Production experiment. The returned Comment in Figure~\ref{fig:comment_and_suggest_hydrogen_production} (resp. Figure~\ref{fig:comment_and_suggest_sugar_beet}) contains examples of the next $n_{\text{LLM}}$ points to evaluate for the Hydrogen Production (resp. Sugar Beet Production), which are then evaluated and added to the dataset.

\clearpage
\subsubsection{$a_{3}$ LLM Comments and Selects $n_{\text{LBO}}$ BO Points}
\begin{figure}[h]
    \centering
    \includegraphics[width=\columnwidth]{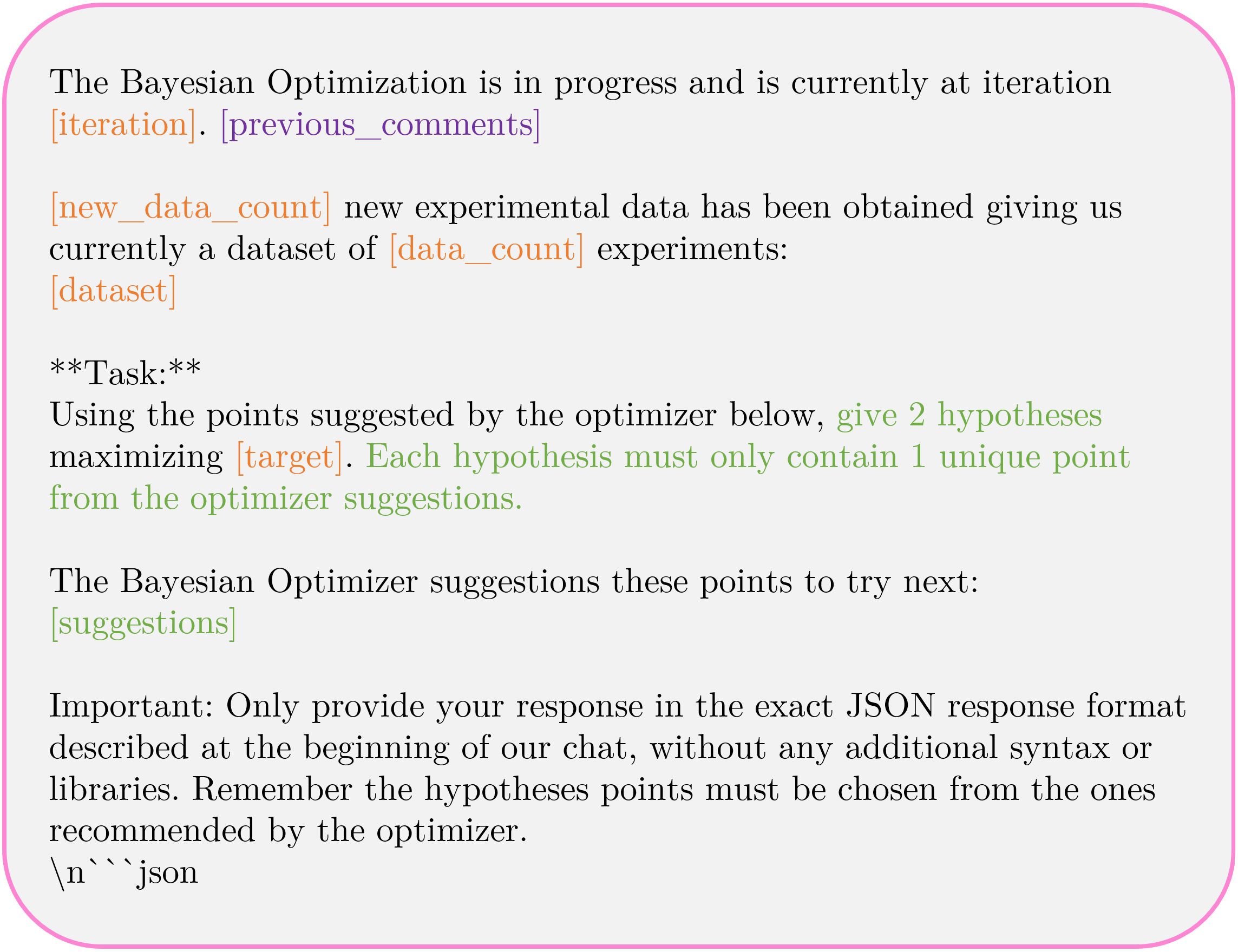}
    \caption{Prompt template tasking the LLM to perform action $a_{3}$ (comment and select the next 2 points to evaluate).}
    \label{fig:comment_and_select_prompt_template}
\end{figure}

The LLM is prompted following the template shown in Figure~\ref{fig:comment_and_select_prompt_template} with the gathered data up to the current iteration, a set of $n_{\text{BO}}=5$ BO-suggested points, and any previous comments. Then, it is tasked to comment on the optimization progress in light of the new data and is constrained to select the $n_{\text{LBO}}=2$ most promising BO candidates that best align with its hypotheses for maximizing the target objective.  

\begin{figure}[!t]
    \centering
    \includegraphics[width=\columnwidth]{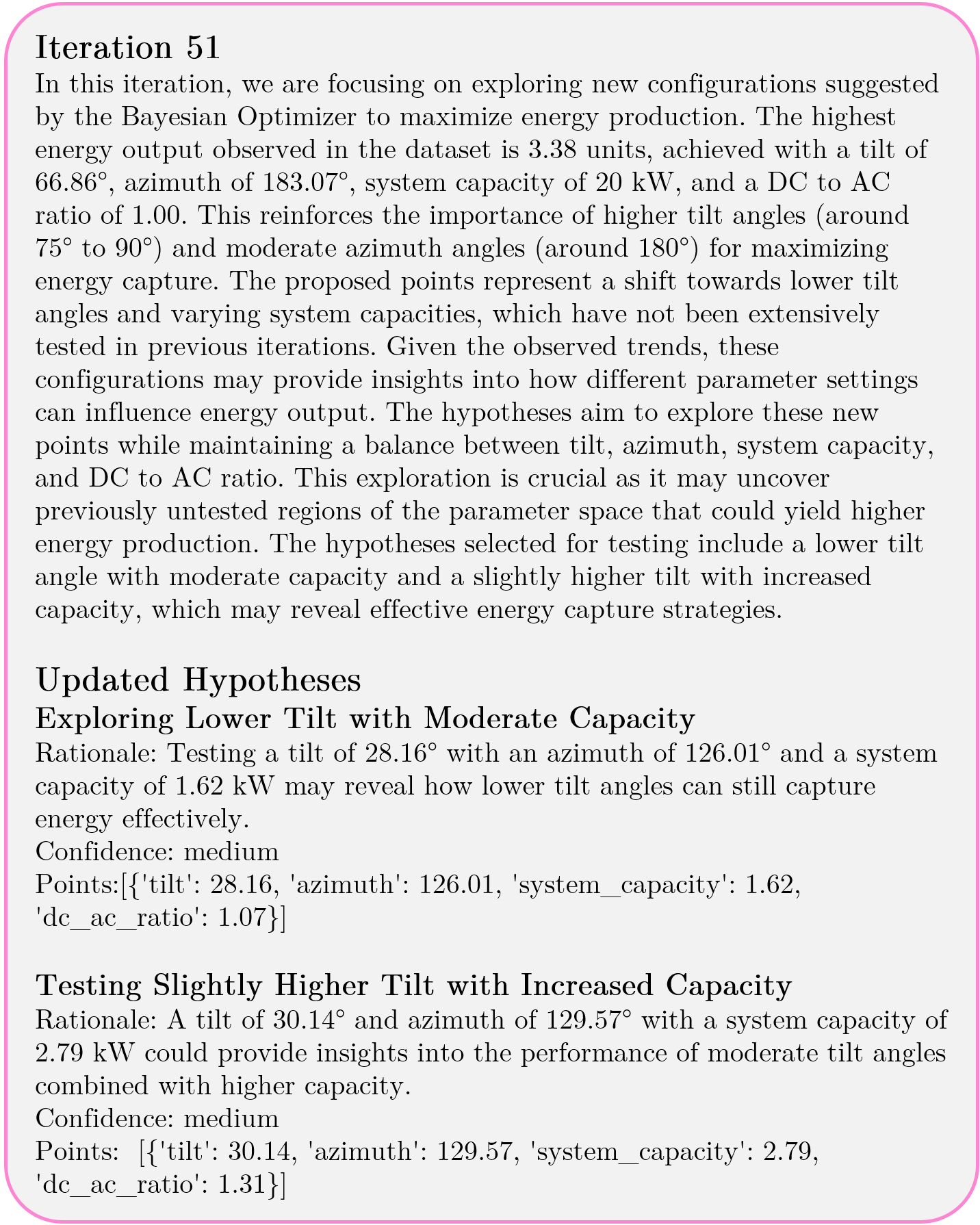}
    \caption{An example of the LLM performing action $a_{3}$ (comment and select the next 2 points to evaluate) on the Solar Energy Production experiment.}
    \label{fig:comment_and_select_solar_energy_production}
\end{figure}

The returned Comment, illustrated in Figure~\ref{fig:comment_and_select_solar_energy_production} for the Solar Energy Production experiment, contains the two selected points that are then evaluated and added to the dataset.

\subsection{Optimization Conclusions and Summary}
After the optimization ends, that is when the sample budget is exhausted, the LLM reflects on its hypotheses to give a brief conclusion to the user before generating a final report summarizing the optimization findings and suggesting further future experiments.

\subsubsection{Hypothesis Evolution}

\begin{figure}[!b]
    \centering
    \includegraphics[width=\columnwidth]{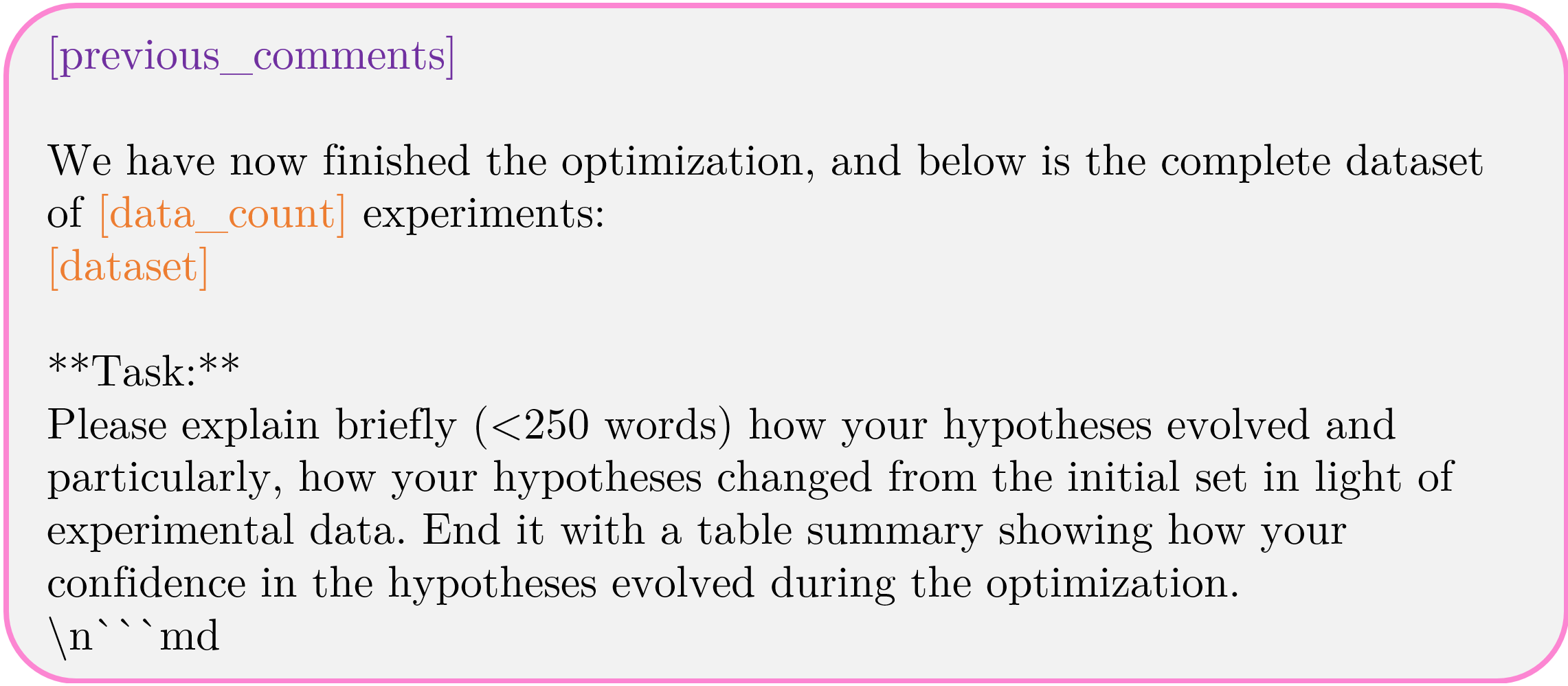}
    \caption{Prompt template tasking the LLM conclude about the experiment.}
    \label{fig:conclusions_prompt_template}
\end{figure}

Based on the entire dataset of the optimization and all its previous comments and hypotheses, the LLM is tasked following the template in Figure~\ref{fig:conclusions_prompt_template} to briefly explain how its hypotheses evolved throughout the optimization in the form of a table that is easily digestible by the user, adhering to the user-centric design of BORA. 

\begin{figure}[!t]
    \centering
    \includegraphics[width=\columnwidth]{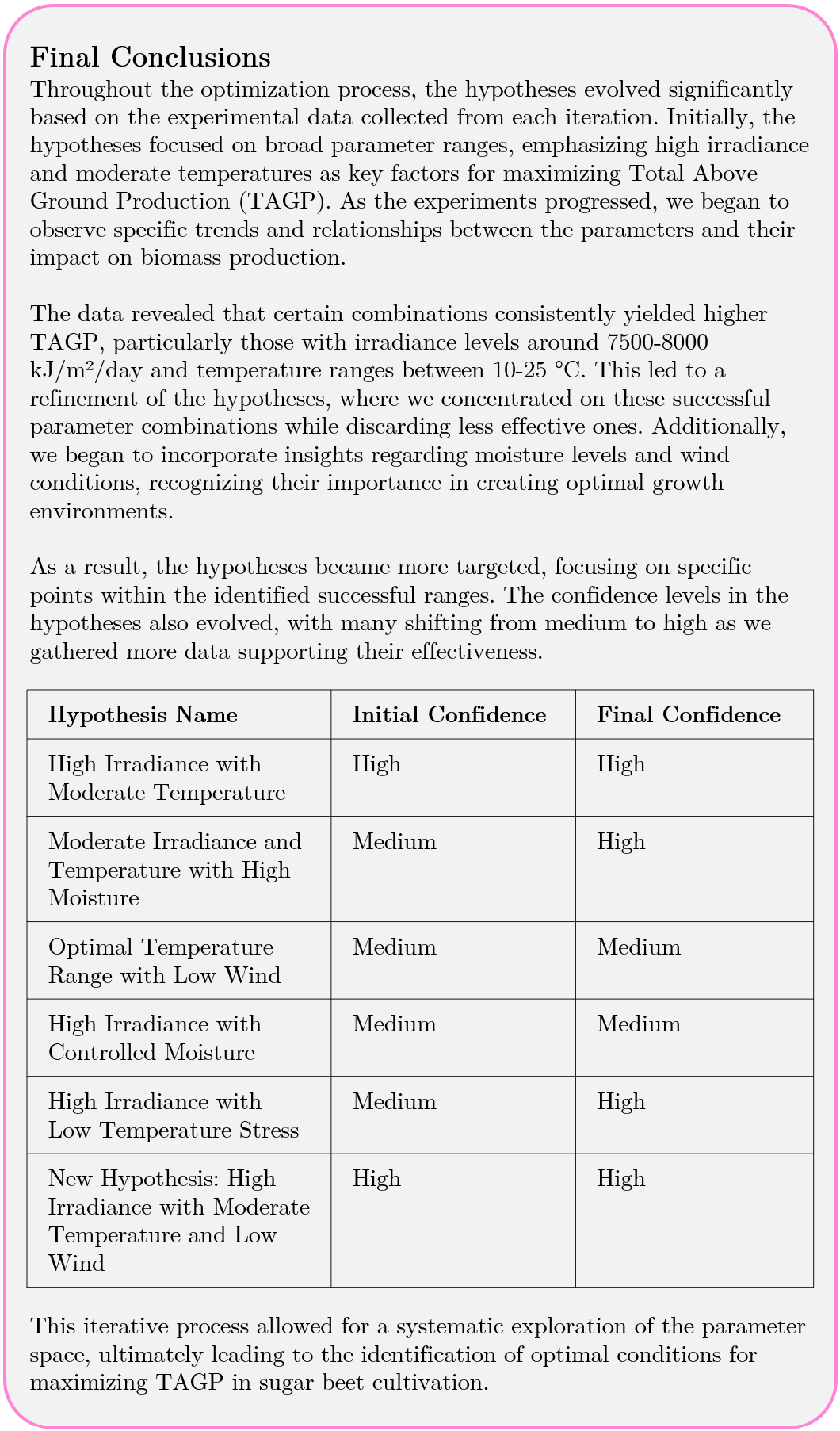}
    \caption{An example of the LLM giving its conclusions about the Sugar Beet Production experiment.}
    \label{fig:conclusions_sugar_beet_production}
\end{figure}

An example of the output of this task is shown in Figure~\ref{fig:conclusions_sugar_beet_production}.

\begin{figure}[!t]
    \centering
    \includegraphics[width=\columnwidth]{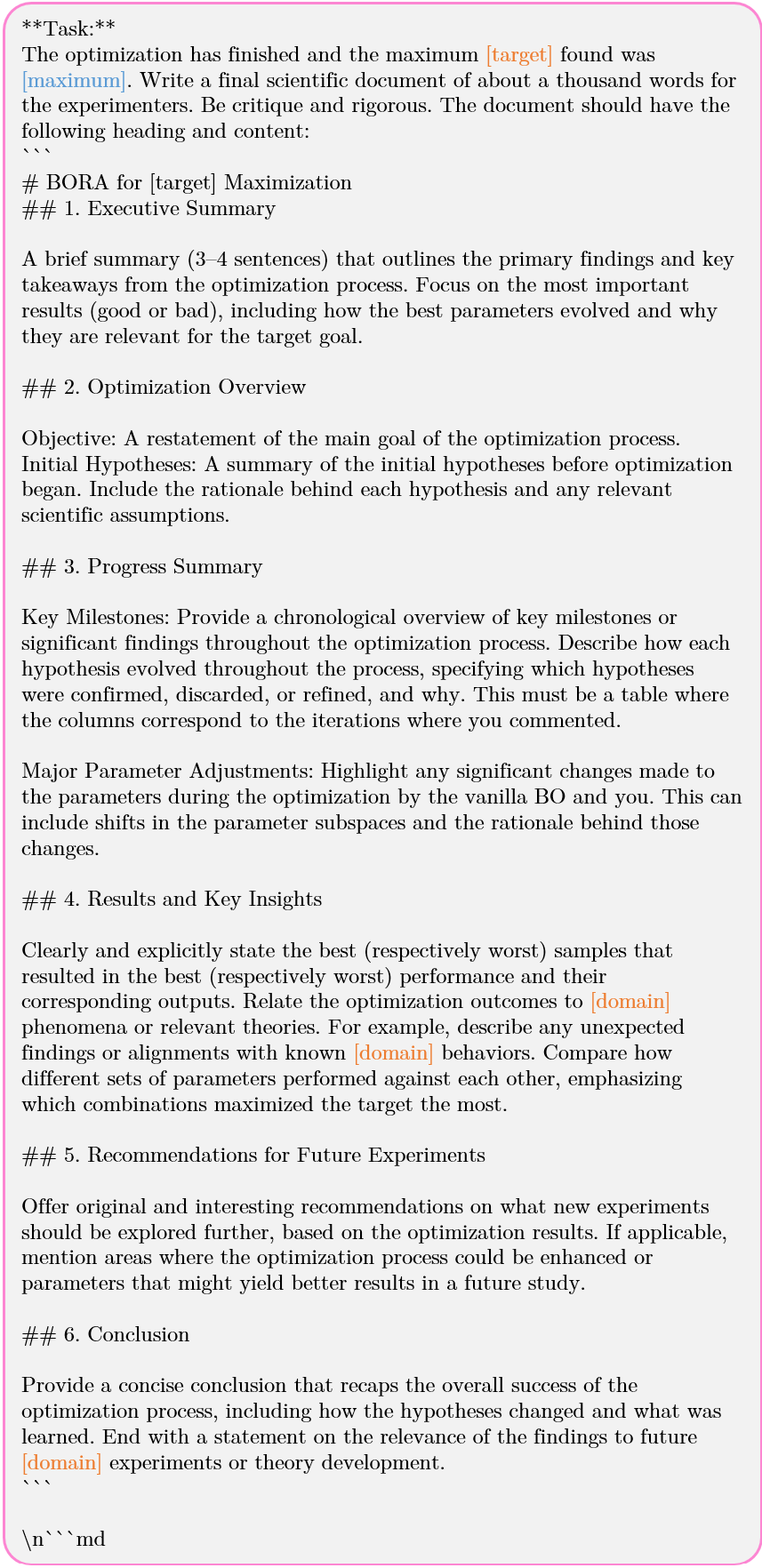}
    \caption{Prompt template tasking the LLM to generate the summary report after optimization concludes.}
    \label{fig:summary_prompt_template}
\end{figure}

\subsubsection{Optimization Summary}
The LLM generates a final summary report to provide users with an overview of the entire optimization process and recommend future experiments. The corresponding prompt, shown in Figure~\ref{fig:summary_prompt_template}, builds on the previous exchanges in the LLM's memory and instructs the LLM to:
\begin{itemize}
    \item Give an overview of the optimization.
    \item Highlight the key results and findings.
    \item Recommend future experiments based on the findings.
\end{itemize}

By leveraging historical context and a structured narrative, the summary ensures that users gain a clear understanding of the optimization’s successes, limitations, and potential next steps.

\clearpage
\subsection{LLM Reflection and Fallback Strategies}
\label{s:reflections}
Although few-shot prompting can enhance the LLM agent outputs, the output quality can be still improved with reflection strategies. This is beneficial for all LLM tasks, whether it is to directly intervene in the optimization process to initialize it or through its actions $a_{2}$ and $a_{3}$, or to generate conclusions and summaries at the end of the optimization. While the self-consistency strategy increases the quality of the LLM output and reduces the likelihood of producing invalid comments, the LLM can still sometimes return invalid Comments during the more direct LLM interventions (Initialization, actions $a_{2}$ and $a_3$). A Comment is valid when:
\begin{itemize}
    \item The \emph{Comment} format is respected.
    \item All hypotheses of the comment are also valid. A valid hypothesis has points that are not duplicates of previously sampled points, are within the bounds of the experiment, and satisfy any existing experimental constraints given in the experiment card.
\end{itemize}
Consequently, we have designed a comment feasibility verification, and in the event of an invalid comment, fallback mechanisms have been developed to allow BORA to proceed with partial or no contribution from the LLM intervention.

\subsubsection{Self-consistency}
For the LLM interventions $a_{2}$ and $a_{3}$, we leverage self-consistency by generating $n=3$ outputs from the same prompt. Each of those outputs reflects a potential viewpoint. These outputs are then systematically consolidated to synthesize a comprehensive and unified response. This consolidation process involves several key steps: (a) reflecting and critiquing each output, (b) evaluating consistency, and resolving discrepancies by cross-referencing them against the optimization dataset. This self-consistency procedure helps maintain coherence and accuracy, which minimizes the risk of biases from a single output while fostering a more robust and nuanced proposition.

\subsubsection{Feasibility Verification and Fallback Mechanisms}
When the LLM is invoked to generate a Comment, the latter is systematically checked to verify that the Comment structure is respected and each hypothesis is valid.

\paragraph{In the Initialization Phase}
In the event of invalid hypotheses, the latter are removed from the comment, resulting in the LLM generating less than $n_{\text{hypotheses}}$ hypotheses; that is, less than $n_{\text{init}}$ initial samples. In such case, a fallback mechanism replaces the missing points with randomly generated points to complete the initial dataset size of $ |\mathcal{D}_{0}| = n_{\text{init}}$ samples.

\paragraph{During LLM Interventions}
For the LLM direct interventions in the optimization process via actions $a_{2}$ and $a_{3}$, in the event that the Comment validation step fails, the LLM is tasked to update its Comment to remedy the validation error, i.e., satisfy the novelty and/or experiment requirements. The LLM is given three attempts, after which the LLM intervention is deemed as failed, which triggers a fallback mechanism. In particular, the LLM intervention is skipped, essentially replacing it with the Vanilla BO action $a_{1}$.

\paragraph{Context Window Limit}
While context window limits were a pronounced limitation when LLMs were first introduced, their capabilities here have increased markedly, resulting in much larger context windows. This is likely to evolve further. To this point, although we have designed a summarizing solution in case the LLM's context window is reached, we have not encountered that scenario during our tests here. This could occur with more complex or longer experiments. Our summarizing mechanism works as follows. We track the number of tokens in the LLM agent's chat history and whether the context window will be exhausted at the subsequent LLM intervention. In such a scenario, instead of giving all the previously gathered data to the LLM as a part of its intervention package, we supply it with the correlation matrix of the data instead. The correlation matrix summarizes the pairwise relationships between the input variables of the problem at hand, considerably reducing the input size for the LLM. In fact, for a dataset with $d$ input variables and $N >> d$ number of samples, the correlation matrix has only $d \times d$ elements instead of the raw dataset with $d \times N$ elements. Although the correlation matrix summarizes the data, the LLM loses the intricate details in the raw data. However, as a result of a condensed data summary, the LLM processes it much faster.

\end{document}